\documentclass{article}

\usepackage[preprint]{neurips_2022}

\usepackage[utf8]{inputenc} 
\usepackage[T1]{fontenc}    
\usepackage{hyperref}       
\usepackage{url}            
\usepackage{booktabs}       
\usepackage{amsfonts}       
\usepackage{nicefrac}       
\usepackage{microtype}      

\usepackage{pifont}
\usepackage{url}
\usepackage[utf8]{inputenc} 
\usepackage[T1]{fontenc}    
\usepackage{url}            
\usepackage{booktabs}       
\usepackage{amsfonts}       
\usepackage{nicefrac}       
\usepackage{microtype}      
\usepackage{times}
\usepackage[table,xcdraw]{xcolor}
\usepackage{color}
\usepackage{soul}
\usepackage[utf8]{inputenc}
\usepackage{graphicx} 
\usepackage{wrapfig}
\usepackage{amssymb}
\usepackage{amsmath}
\usepackage{multirow}
\usepackage{float}
\usepackage{subcaption}
\usepackage{arydshln}

\usepackage{times}
\usepackage{epsfig}
\usepackage{graphicx}
\usepackage{amsmath}
\usepackage{amssymb}
\usepackage{textcomp}
\usepackage{wrapfig}

\usepackage{bbding}
\usepackage{pifont}
\usepackage{wasysym}
\usepackage{amssymb}
\usepackage{arydshln}
\usepackage{cases}
\usepackage[ruled,vlined,linesnumbered]{algorithm2e}

\title{
CodeRL: Mastering Code Generation through Pretrained Models and Deep Reinforcement Learning
}

\author{%
  Hung Le\thanks{Equal contribution.},  Yue Wang\footnotemark[1],  Akhilesh Deepak Gotmare, Silvio Savarese, Steven C.H. Hoi
  \thanks{Corresponding authors: \texttt{\small \{hungle, shoi\}@salesforce.com}}
  \\
  Salesforce Research\\
  \url{https://github.com/salesforce/CodeRL}\vspace{-0.3in}
}

\begin{document}

\maketitle

\begin{abstract}

Program synthesis or code generation aims to generate a program that satisfies a problem specification. Recent approaches using large-scale pretrained language models (LMs) have shown promising results, yet they have some critical limitations. In particular, they often follow a standard supervised fine-tuning procedure to train a
code generation model only from the pairs of natural-language problem descriptions and ground-truth programs. Such paradigm largely ignores some important but potentially useful signals in the problem specification such as unit tests, 
which thus often results in poor performance when solving complex unseen coding tasks. To address the limitations, we propose ``CodeRL'', a new framework for program synthesis tasks through pretrained LMs and deep reinforcement learning (RL). Specifically, during training, we treat the code-generating LM as an actor network, and introduce a critic network that is trained to predict the functional correctness of generated programs and provide dense feedback signals to the actor. During inference, we introduce a new generation procedure with a critical sampling strategy that allows a model to automatically regenerate programs based on feedback from example unit tests and critic scores. For the model backbones, we extended the encoder-decoder architecture of CodeT5 with enhanced learning objectives, larger model sizes, and better pretraining data. Our method not only achieves new SOTA results on the challenging APPS benchmark, but also shows strong zero-shot transfer capability with new SOTA results on the simpler MBPP benchmark.

\end{abstract}

\begin{figure}[htbp]
	\centering
	\resizebox{1.0\textwidth}{!} {
	\includegraphics{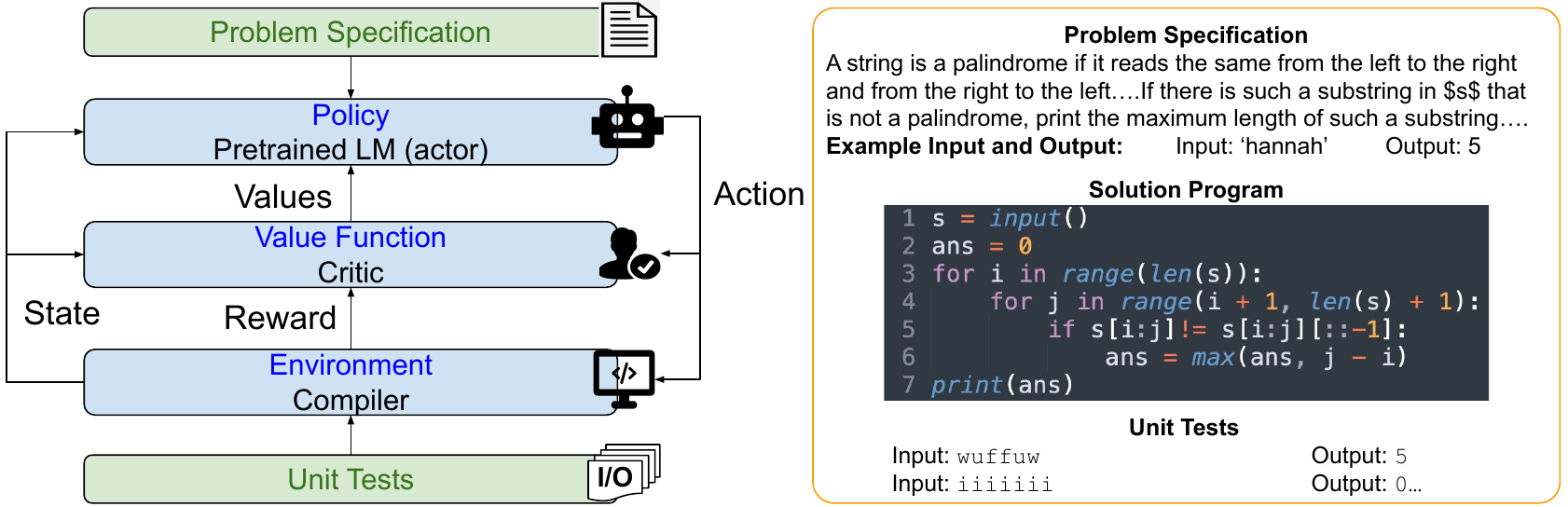}
	}
	\caption{
	\textbf{An example program synthesis task (Right):}
	Each task is defined by a problem specification in natural language, often containing example input and output pairs. 
	The expected output is a program to be checked for functional correctness against some unit tests. 
	\textbf{A high-level overview of our CodeRL framework for program synthesis (Left):} 
	we treat a pretrained code language model (LM) as a stochastic policy, code generations as actions, and rewards can be estimated based on the unit test results of output programs from the compiler (environment).
	}
	\label{fig:example_program}
\end{figure}

\newpage
\tableofcontents
\newpage

\section{Introduction}

Program synthesis or code generation is the task of designing and building an executable computer program that satisfies a problem specification (see Figure \ref{fig:example_program}, right, for an example). 
Program synthesis research has gained much attention due to its significant impacts on the software industry, including better productivity and accessibility of programming jobs and education. 
Developing an AI model that can automatically generate programs based on human requirements can dramatically transform programming tools and the way we work with them. 

Recent attempts employ deep learning methods, specifically Transformer-based pretrained language models (LMs) \citep{vaswani2017attention, brown2020language}, which were originally intended for natural language learning tasks, to generate unique computer programs. 
These approaches \citep{ hendrycksapps2021, chen2021evaluating, austin2021program} consider program synthesis as a sequence-to-sequence task, which receives input sequence as problem specification in natural language and generates a sequence of codes as the output program.
While these models achieve promising results, especially in basic programming tasks \citep{chen2021evaluating, austin2021program}, we observe that they still fail to generate codes to solve complex problems such as those at programming competitions \citep{hendrycksapps2021, li2022competition}.

There are two main limitations. 
First, current models are trained using a conventional next-token prediction (NTP) objective which maximizes the next ground-truth token likelihood. 
As noted in NLP domains \citep{bengio2015scheduled, DBLP:journals/corr/RanzatoCAZ15}, training models only with next-token prediction objective in a "teacher-forcing" manner often leads to accumulating errors during test time when tokens are generated by conditioning on previously sampled tokens, not the ground-truth tokens.
This issue becomes more serious in the domain of program synthesis, where token-matching scores such as BLEU \citep{papineni2002bleu, ren2020codebleu} are more appropriate in partial program synthesis tasks (i.e. code completion) \citep{csn} but have failed to measure the functional correctness of complete programs \citep{hendrycksapps2021, chen2021evaluating}. 
Training only with NTP objective is hence, not ideal to tackle full program generation to solve programming problems. 

Secondly, current models fail to utilize the potential meaningful signals from unit tests, which directly determine the model performance by the functional correctness of programs. 
Current approaches neglect this important signal during model optimization as well as generation procedure.
During optimization, unit tests could be factored into learning objectives to match the final goal of generating semantically correct programs. 
During inference, since unit tests are often parts of problem description (i.e. example unit tests), they are potentially powerful to further improve output programs. 
Related approaches such as \citep{li2022competition} use example unit tests to filter and rank final output programs.
While this method naturally selects better program candidates, it does not allow models to improve the programs based on the initial (example) unit test results.

To address the above issues, we introduce ``CodeRL'', a new framework to improve pretrained LMs for program synthesis tasks through deep reinforcement learning (see Figure \ref{fig:example_program}, left, and Section \ref{sec:coderl} for more details). 
Specifically, we propose a training strategy that optimizes pretrained LMs for program synthesis tasks in an actor-critic RL approach \citep{konda1999actor, sutton1999policy}.
We treat the pretrained LM as an actor network and synthetically sample sequences from this actor, including both correct and incorrect programs.
These program samples are passed to a critic model which is trained as an error predictor to assess the functional correctness of these samples.
We use the token-level hidden states extracted from the learned critic model to estimate the values/scores of output tokens of these synthetic samples. 
The actor network is then finetuned on these synthetic samples weighted by their critic scores. 

During inference, as part of the CodeRL framework, we introduce a new generation procedure that systematically exploits example unit test signals to allow models to further improve programs. 
Firstly, for samples that pass the example unit tests, we employ the critic model to filter and select sub-sequences.
These sub-sequences are utilized as ``seeds'' that condition the model to resample new tokens and obtain new output programs. 
Secondly, among failed programs, the critic selects top programs based on their likelihood of passing unit tests. 
These program candidates are concatenated with the error information received from a compiler and passed to a program repair module.
This generation procedure enables a dual strategy to automatically refine and repair output programs based on their functional correctness during test time. 


Together with CodeRL, we extend CodeT5 as a foundation model with improved pretraining strategies, including better pretraining objectives, larger model sizes, and massive pretraining data. 
Our comprehensive experiments (Section \ref{sec:experiments}) show that our models can achieve SOTA performance on the challenging APPS benchmark \citep{hendrycksapps2021}.
Specifically, our models reach more than 2\% \emph{pass@1}, 6\% \emph{pass@5}, and 20\% \emph{pass@1000}. 
Since our RL method is model-agnostic, we also apply it to various large-scale models and achieve consistent performance gains. 
We further test its zero-shot transfer ability on a simpler MBPP benchmark~\citep{austin2021program}, where it sets a new SOTA result of 63.0\% \emph{pass@80} over a finetuned GPT-137B's 61.4\%.
We perform qualitative analysis to understand the problems that the model succeeds or fails to solve. 
Finally, we release the improved CodeT5-large ($770$M) model which outperforms many pretrained LMs of much larger sizes.


\section{Related Work}
\subsection{Program Synthesis}
Program synthesis tasks can date back as early as the early adoption of machine learning research \citep{waldinger1969prow, manna1971toward}.
Earlier tasks include problem specifications in the form of input-output (IO) examples \citep{summers1977methodology, gulwani2012spreadsheet} and synthesis methods are limited to probabilistic approaches \citep{liang2010learning} or simple programming concepts \citep{joulin2015inferring, kurach2015neural}. 
As deep learning methods became popular, later approaches adopt neural models to induce output programs, assuming an inductive bias given a sufficient number of program samples \citep{parisotto2016neuro, balog2016deepcoder, devlin2017robustfill}. 

More recently, we witnessed the emergence of program synthesis tasks in which output programs are extended to general-purpose programming languages \citep{yin-neubig-2017-syntactic, xu2018sqlnet, chen2021evaluating} and program specifications are fully described in natural English text \citep{hendrycksapps2021, austin2021program, poesia2022synchromesh}.
These extensions have encouraged a rising number of applications of pretrained language models (LMs) to program synthesis to exploit the contextual representations learned from massive data of codes and natural languages \citep{feng-etal-2020-codebert, clement-etal-2020-pymt5, codet5, gpt-j, ChenVarCLR2022}.
\cite{nijkamp2022conversational} proposed a conversational program synthesis approach with large pretrained language models.
Despite impressive results in basic programming problems and initial commercial deployment\footnote{\url{https://copilot.github.com/}}, existing models still perform poorly against complex problems such as those from programming competitions on Codeforces \citep{hendrycksapps2021, li2022competition}. 

\paragraph{Program Synthesis in Visual Context.}
Another related line of research is program synthesis in computer vision domains such as images and videos. 
Early papers such as \citep{kulkarni2015deep, yang2015weakly} introduce inverse graphics networks to infer visual properties such as pose, shape, and lighting, of visual objects. 
\cite{Wu_2017_CVPR, liu2018learning, ellis2018learning} study the problem of image rendering, which transforms an image to structured and compact representations, i.e. \emph{scene programs}. 
\cite{tian2018learning} extends the prior work to render 3D shapes from images through \emph{shape programs}, containing features to capture geometric and structural priors. 
\cite{ganin2018synthesizing} introduces an RL-based approach to render realistic images through high-level graphics programs.
\cite{pmlr-v80-sun18a}introduces program synthesis from demonstration synthetic videos to summarize the behaviors of the objects in the videos. 

While this line of research has remarkable impacts on applications such as image/video editing, captioning, and extrapolating, these approaches are limited to programs of domain-specific languages defined for visual objects. 
For instance, in \citep{pmlr-v80-sun18a}, programming language contains basic functions for object perception, action, and control flows.
In our work, we focus on program synthesis from natural language problem specifications and the output programs are in general-purpose languages such as Python. 
This type of programming task can range from basic programming problems to competition-level programming tasks that require a high level of problem-solving skills. 

\subsection{Reinforcement Learning for Sequence Generation}
Related to the program synthesis tasks are research domains of sequence generation, in which RL approaches have demonstrated remarkable achievements. 
In these domains, RL approaches are used to exploit signals from non-differentiable metrics of the task at hand.
Earlier work such as \citep{DBLP:journals/corr/RanzatoCAZ15} adopts this strategy with REINFORCE algorithm \citep{williams1992simple} to directly optimize models for sequence-based test metrics such as BLEU \citep{papineni2002bleu} and ROUGE \citep{lin2004rouge} scores for translation models. 
In the same domain, \cite{bahdanau2016actor} introduced an actor-critic framework \citep{sutton1984temporal, konda1999actor}.
In visual captioning domains, \cite{rennie2017self, wang2018video} proposed to use RL to optimize image captioning models using variants of CIDEr scores \citep{vedantam2015cider}. 
Alternatively, \cite{ren2017deep} derived a new goal-oriented return estimate using visual-semantic embedding. 
\citet{johnson2017inferring, trivedi2021learning} introduce program generation as an auxiliary task to learn interpretable policies in question-answering and synthetic navigation tasks. 

Different from prior domains, in program synthesis, \citet{austin2021program, chen2021evaluating, li2022competition} demonstrated very low correlation between token-based similarity metrics and functional correctness of programs.
Hence, it is not trivial to define an appropriate optimization goal in this domain.
We propose to exploit unit test signals, which directly exhibit functional correctness of programs, during both - model optimization and test-time generation stages. 
More related to our work are RL-based program synthesis \citep{guu-etal-2017-language, bunel2018leveraging, liang2018memory, zhong2018seqsql} and execution-guided synthesis approaches \citep{ellis2019write, chen2021latent}.
However, these are limited to programming languages defined within a specific application domain only.





\subsection{Program Completion}

Related to our work is the research of automatic program completion or code completion.
Code completion aims to generate programs conditioned on partial codes (e.g. function signatures, code with blank gaps) and the output programs are often short snippets as potential code suggestions. 
Early work such as \citep{robbes2008program, bruch2009learning} shows that sufficient program samples and prior program history can facilitate better code completion systems in terms of the relevance of code suggestions. 
\citet{raychev2014code, white2015toward} introduce deep learning-based approaches by considering the tasks as an NLP problem of predicting probabilities of tokens or sentences using neural language models. 
\citet{svyatkovskiy2021fast, guo2021learning} improve code completion systems with a reranking strategy to select better program candidates and with structured predictions to generate more syntactically correct programs.  
Recent work such as \citep{clement-etal-2020-pymt5, svyatkovskiy2020intellicode} adopt pretrained language models to exploit the learned representations from large source code data and \citep{aye2021learning} tackles real-world code completion. 

Compared to code completion, program synthesis requires systems to generate complete programs from scratch and these programs are typically evaluated by their functional correctness through some unit tests \citep{hendrycksapps2021, li2022competition}. 
In this work, while we focus on program synthesis from natural problem descriptions, we adopt a similar strategy to code completion in our generation procedure to improve output programs. 



\section{CodeRL}
\label{sec:coderl}
\subsection{Program Synthesis Task}
Following a sequence-to-sequence approach, the program synthesis task contains a problem description as an input sequence $D$ and an output sequence of program $\hat{W}=(\hat{w}_1, ...,\hat{w}_T), \hat{w}_t \in \mathcal{V}$ \footnote{For simplicity, we use $T$ as the notation of sequence length for all sequences which can actually be variable.} that can solve the problem.
The output at each decoding step $t$ is a distribution over the vocabulary $\mathcal{V}$, computed by the softmax function $\hat{w}_t \sim \mathrm{softmax}(\mathrm{Linear}(s_{t}))$ where $s_{t}$ is the contextual hidden state at decoding step $t$.
Conventionally, during train time, model parameters, $\theta$, are learned by maximizing the likelihood of the ground-truth reference programs. Denoting $W=(w_1,...w_T)$ as the ground-truth program, the objective is to minimize the cross-entropy loss: 
\begin{align}
    \mathcal{L}_{ce}(\theta) = - \sum_t \log p_\theta(W|D) = - \sum_t \log [p_\theta (w_t | w_{1:t-1}, D)]
\end{align}
where the conditional probability $p_\theta$ is parameterized following the above softmax function. 
During test time, models generate sequences of programs by autoregressively sampling token $\hat{w}_t$ from the distribution $p_\theta (.| \hat{w}_{1:t-1}, D)$.
Models are evaluated against unit tests corresponding to the problem.
Each test includes a pair of input and ground-truth output. 
In real-world program synthesis tasks \citep{hendrycksapps2021}, \emph{example unit tests} are often given as parts of the problem specification.

\subsection{Pretraining Language Models on Code}
\label{subsec:pretrain_lm}

We adopt Transformer models as the backbone of our program synthesis systems. 
Specifically, this paper extends the CodeT5 model \citep{codet5} as a foundation model for CodeRL.

\paragraph{CodeT5.}
CodeT5~\citep{codet5} is a multi-lingual code-aware language model pretrained on large-scale source code corpora curated from Github. 
With a unified encoder-decoder architecture, CodeT5 achieves state-of-the-art performance in  a wide range of code intelligence tasks in the CodeXGLUE benchmark~\citep{codexglue}  including both code understanding and generation tasks. 

\paragraph{Improving Pretraining Data.}
We enlarge the Python pretraining dataset using the recently released large-scale Github Code dataset\footnote{\url{https://huggingface.co/datasets/lvwerra/github-code}}. 
We have compiled public, non-personal information from GitHub consisting of permissively licensed Python code
(e.g. “mit”, “apache-2”, “bsd-3-clause”, “bsd-2- 126 clause”, “cc0-1.0”, “unlicense”, “isc”). The resulting Python dataset (GCPY) has 10.5B tokens and is 10x larger than the CodeSearchNet (CSN) corpus \citep{csn} used in the original CodeT5 \citep{codet5}.
%

\paragraph{Improving Pretraining Objective.}
While pretraining tasks in CodeT5 like masked span prediction (MSP) benefit code understanding tasks, they have a large discrepancy with program synthesis objectives.
To mitigate this gap, we introduce a pretraining task of next-token prediction (NTP) into CodeT5. Specifically, we uniformly sample a pivot location for each code sample, then pass the content preceding the pivot to the encoder and remaining to the decoder.
To control the length of input and output sequences, we restrict the pivot within 10\% to 90\% of the original sequence.  
\subsection{Program Synthesis as an RL Problem}
\label{subsec:rl_finetune}

\begin{figure}[t]
	\centering
	\resizebox{1.0\textwidth}{!} {
	\includegraphics{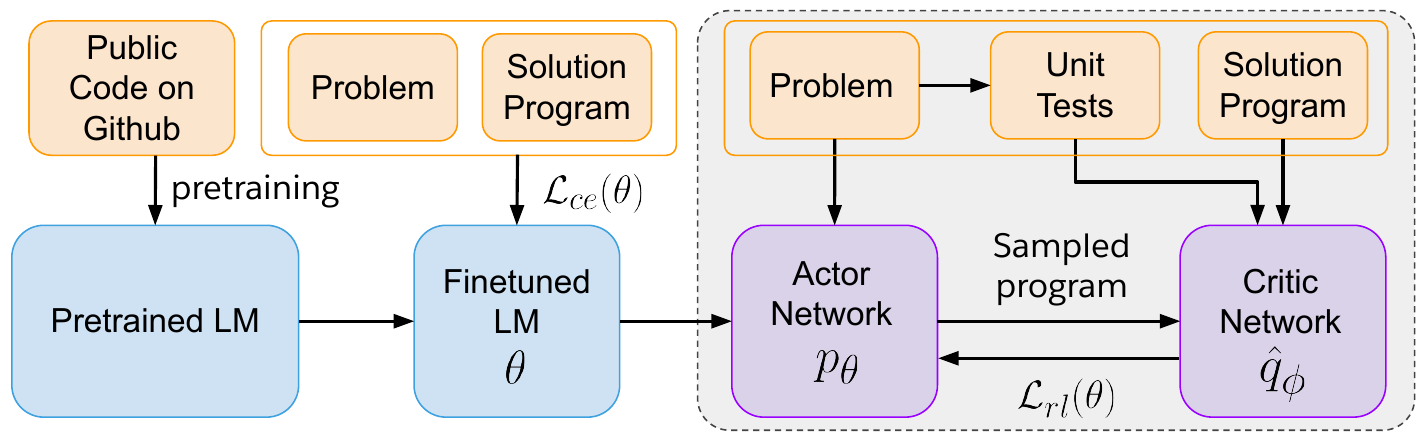}
	}
	\caption{
	\textbf{Overview of our actor-critic framework to optimize pretrained LMs for program synthesis:}
	We treat the LM as an actor network and sample synthetic samples from this actor.
	Another neural network is trained as a critic model to evaluate these synthetic samples based on their probabilities of passing unit tests. 
	The returns are estimated based on critic scores and finally factored into the learning objective $\mathcal{L}_{rl}$ to finetune the actor LM network using synthetic samples. 
	}
	\label{fig:rl_approach}
\end{figure}

We propose to formulate the Program Synthesis as an RL problem (see Figure \ref{fig:example_program}, left) and apply an actor-critic RL approach to improve the performance of a pretrained LM by exploiting the unit test signals in both model optimization (see Figure \ref{fig:rl_approach}) and generation procedure (see Figure \ref{fig:critic_sampling}). 

More formally, we can view the learned parameters of an LM model, $\theta$ as a stochastic \emph{policy}, which decides an \emph{action} as the prediction of each token.
Following each action, an LM model updates its hidden state representations which are used by the policy to determine the next action in the next decoding step. 
At the end of the generation episode (i.e. an \emph{<endoftext>} token is observed), the LM model receives a \emph{return} $r$ measured by the functional correctness of the generated program. 
The goal of RL finetuning is to minimize the expected return: 
\begin{align}
    \mathcal{L}_{rl}(\theta) = - \mathbb{E}_{W^s \sim p_\theta} [r(W^s)]
\end{align}
where $W^s=(w^s_1, ..., w^s_T)$ is a synthetic sample in which each token $w^s_t$ is sampled by the LM model at decoding time step $t$.
Following the REINFORCE algorithm \citep{williams1992simple, sutton2018reinforcement} and policy gradient theorem \citep{sutton1999policy} we can define an estimate  of the gradient $\nabla_\theta L(\theta)$ of the non-differentiable return $r$ as: 
\begin{align}
    \nabla_\theta \mathcal{L}_{rl}(\theta) &\approx -\mathbb{E}_{W^s \sim p_\theta} [r(W^s) \nabla_\theta \log p_\theta(W^s|D)] \nonumber \\
    &\approx -\mathbb{E}_{W^s \sim p_\theta} [r(W^s)  \sum_t \nabla_\theta \log p_\theta(w^s_t|w^s_{1:t-1}, D)] \label{eq:rl_loss}
\end{align}
\subsubsection{Defining Return by Unit Test Signals}
For each sample sequence $W^s$, the return $r$ can be defined heuristically by checking its functional correctness. 
We pass generated programs together with the corresponding unit tests to a compiler.
From the outputs of the tests, we can determine the return $r$: 
\begin{numcases}{r(W^s) =}
$-1.0$   & \quad \text{, if } $W^s$ \text{ cannot be compiled (i.e. compile error)} \label{eq:return1}\\
$-0.6$  & \quad \text{, if } $W^s$ \text{ cannot be executed with unit tests (i.e. runtime error)} \\
$-0.3$  & \quad \text{, if } $W^s$ \text{ failed any unit test} \\
$+1.0$  & \quad \text{, if } $W^s$ \text{ passed all unit tests}
\label{eq:return2}
\end{numcases}
However, in related domains such as text-to-SQL research \citep{zhong2018seqsql, xu2018sqlnet}, we note that this approach to estimate returns can lead to unstable model training with high variance of the gradient estimate following Eq.~(\ref{eq:rl_loss}) with mini-batches in training. 

\begin{figure}[t]
	\centering
	\resizebox{1.0\textwidth}{!} {
	\includegraphics{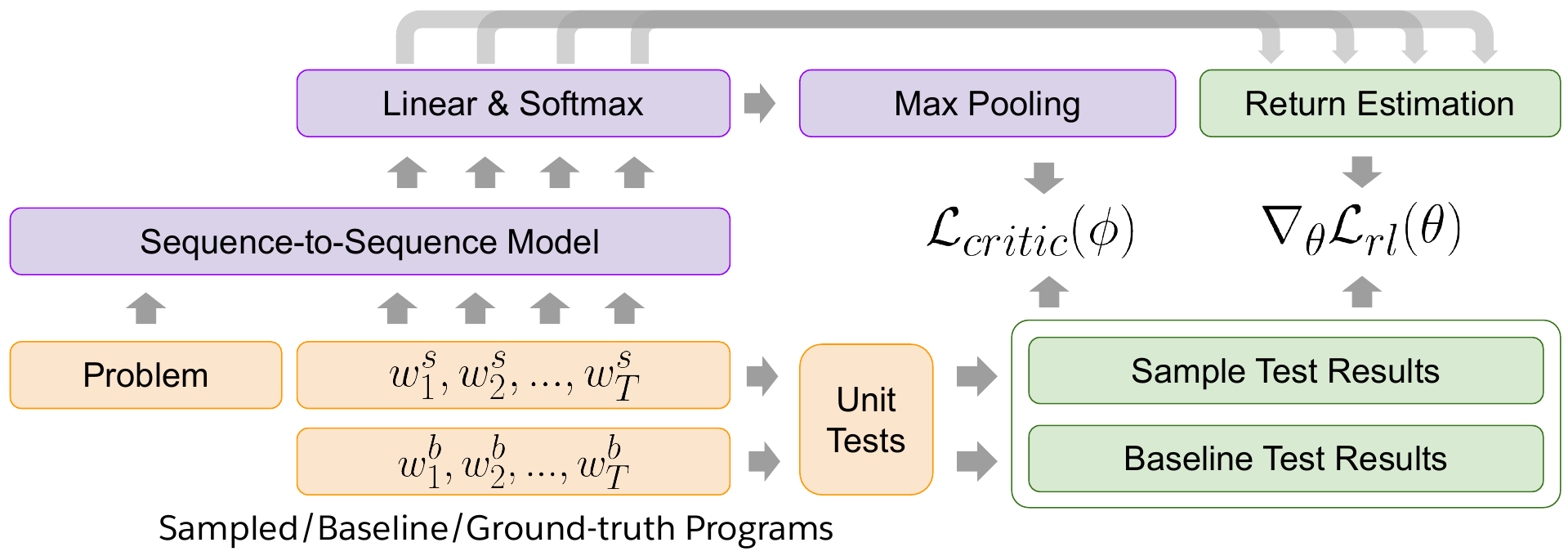}
	}
	\caption{
	\textbf{Overview of our critic model:}
	The critic model is learned as an error predictor.
	The model receives problem specifications and programs as input sequences. 
	For each program, the model is trained to predict one of four possible test outcomes: 
	$\{\mathrm{CompileError, RuntimeError, FailedTest, PassedTest}\}$.
	The learned hidden state representations from the critic are then used to estimate returns of synthetic samples to finetune the actor network. 
	To improve and stabilize the training process, baseline programs are considered and relative returns are factored into the loss function to optimize the actor network. 
	}
	\label{fig:critic_model}
\end{figure}

\subsubsection{Return with a Baseline}
In order to alleviate this variance, we adopt a ``baseline'' \citep{sutton2018reinforcement}. 
Specifically, we use a greedy decoding strategy as a baseline and any generated samples that outperform this baseline are given positive return estimation, and negative return estimation otherwise. 
This relative normalization technique allows models to explore imperfect programs, as long as their returns are better than the baseline's.
Given a training sample, we denote the return of the baseline $r(W^b)$ and 
the expected gradient is computed as: 
\begin{align}
    \nabla_\theta \mathcal{L}_{rl}(\theta) \approx -\mathbb{E}_{W^s \sim p_\theta} [(r(W^s)-r(W^b))  \sum_t \nabla_\theta \log p_\theta(w^s_t|w^s_{1:t-1}, D)]
    \label{eq:loss_rl}
\end{align}
Note that at each decoding step $t$, our greedy decoding baseline is independent from the action $w^s_t$ and hence, the expected gradient term $\nabla_\theta \mathcal{L}_{rl}(\theta)$ from Eq. (\ref{eq:rl_loss}) remains the same in Eq. (\ref{eq:loss_rl}).


\subsubsection{Intermediate Return by Critic as Error Predictor}
We observe that the above gradient estimate is only based on a final return at the end of the decoding process. 
However, programs often follow fixed syntactical rules in which a single token such as an additional white-space can render a program erroneous. 
Therefore, Eq. (\ref{eq:loss_rl}) becomes too restrictive. 
A straightforward solution is to use token-based similarity scores \citep{papineni2002bleu, ren2020codebleu}) between each subsequence $W^s_{1:t}$ and the ground truth. 
However, code matching is not an ideal return measure due to its poor correlation with program correctness \citep{hendrycksapps2021, chen2021evaluating, austin2021program} which can only be measured against fully complete programs. 

Alternatively, we introduce a \emph{critic} model.
Figure \ref{fig:critic_model} shows an overview of our critic model.
The critic model is parameterized as a neural network with parameters $\phi$ that receives inputs as the problem description $D$ and a sampled program $W^s$ = $\{w^s_{1}, \dots, w^s_{T} \}$.
The critic is trained to infer the unit test outcome; one of $\{\mathrm{CompileError, RuntimeError, FailedTest, PassedTest}\}$ as described in the return definitions in Eq. (\ref{eq:return1}) to (\ref{eq:return2}). The training objective of the critic $\phi$ can be expressed as: 
\begin{align}
    \mathcal{L}_{critic}(\phi) &= -\log p_\phi (u| W^s, D) \label{eq:critic_obj}
\end{align}
where $u$ denotes the ground-truth unit test outcome given by the compiler after passing $W^s$ to the unit tests corresponding to the problem.
We use Transformer models of smaller sizes than the actor model as the base architecture for the critic model. The contextual hidden states of the program tokens ($h_1, \dots, h_T$) obtained from the critic model decoder are max-pooled along the sequence length dimension $h^\mathrm{pool} = \mathrm{Pooling}(h_1, \dots, h_T)$. 
The critic's prediction on the unit test outcome is computed as $\hat{u} = \mathrm{softmax}(\mathrm{Linear}(h^\mathrm{pool}))$.

Given a learned critic, we use the probability distribution $\hat{v}_t = \mathrm{softmax}(\mathrm{Linear}(h_{t}))$ to estimate the token-level value $\hat{q}$ of $w^s_t$ \emph{in relation to} the ground-truth unit test output (note that we use the token level contextual representation $h_t$ here, before the pooling operation). Specifically, $\hat{q}_\phi(w^s_t) = \hat{v}_{t}[u]$ where $\hat{v}[.]$ denotes the probability of a specific unit test outcome from the four possible ones. We use this estimate to train the actor LM model with intermediate returns:
\begin{align}
    \nabla_\theta \mathcal{L}_{rl}(\theta) \approx -\mathbb{E}_{W^s \sim p_\theta} [(r(W^s)-r(W^b))  \sum_t \hat{q}_\phi(w^s_t) \nabla_\theta \log p_\theta(w^s_t|w^s_{1:t-1}, D)]
    \label{eq:rl_loss_final}
\end{align}


Note that since our critic model is applied in a supervised learning environment with available ground truth, we also use the training samples of perfect output programs $W$ and assign them with the default test outcome $u = $ \emph{PassedTest} to train the critic.

\begin{figure}[t]
	\centering
	\resizebox{1.0\textwidth}{!} {
	\includegraphics{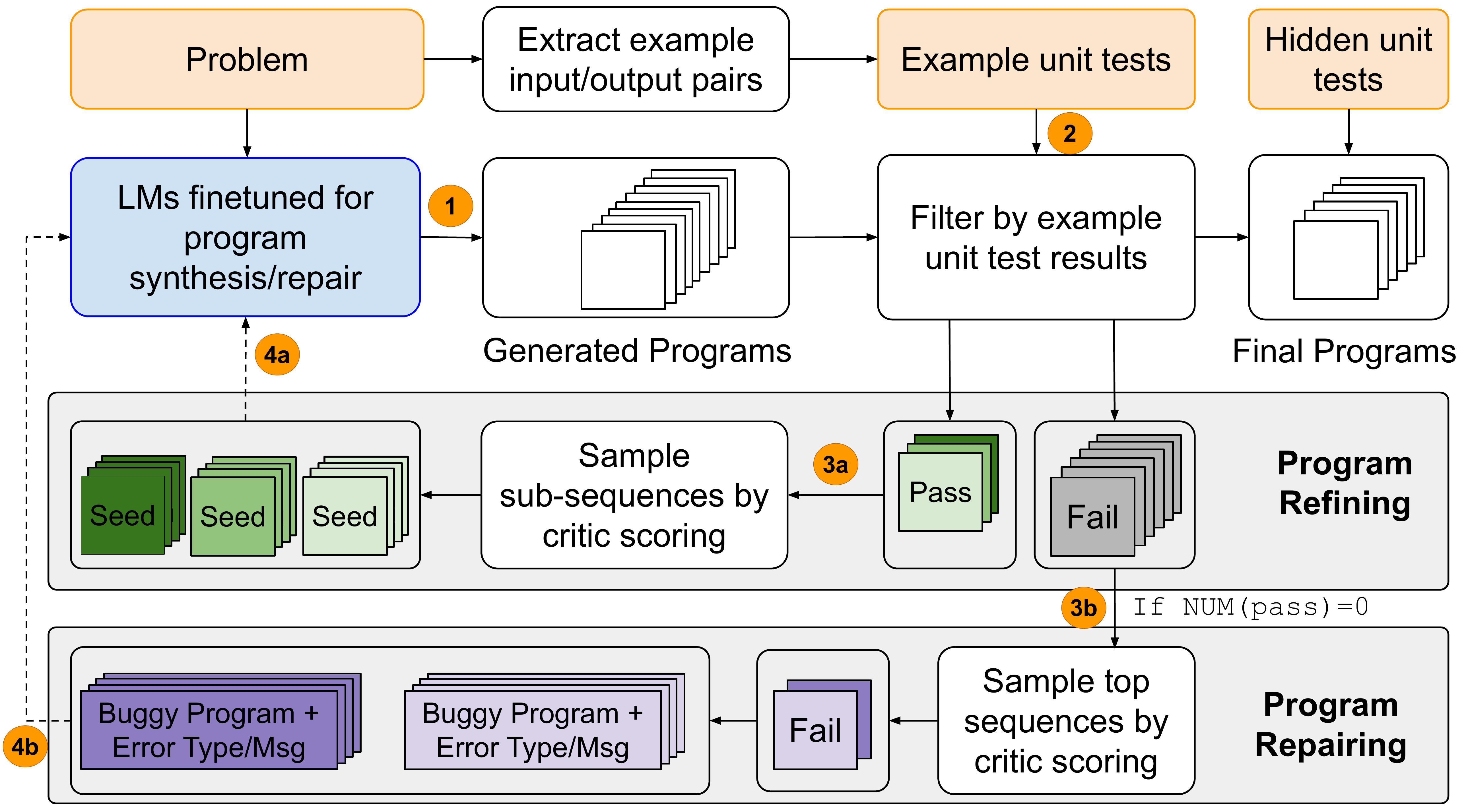}
	}
	\caption{
	\textbf{Overview of our Critic Sampling approach for program synthesis during inference:} 
	(1) For each test problem, we first use finetuned LM to generate a set of solution programs. 
	(2) From the problem description, we extract example unit tests and test against generated solutions. 
	(3a) If there are any passed solutions, we pass them to the critic model to sample sub-sequences. 
	(4a) The sub-sequences from (3a) are used as seed sequences to condition the LM to regenerate solution programs, repeating the steps from (1) onward.
	(3b) If there are no passed solutions from (2), we pass current solutions to the critic model to select the best candidates.
	(4b) These candidates from (3b) are passed to a program repair model, together with their error types and compiler error messages. 
	We repeat the same steps as (2) onward. 
	Dotted lines indicate optional processes that apply during program refining or repairing.
	}
	\label{fig:critic_sampling}
\end{figure}

\subsubsection{Generating Programs with Example Unit Tests and Critic}
We leverage the unit tests provided in the input problem description to improve the generation procedure during inference too (see Figure \ref{fig:critic_sampling} for an overview and Appendix \ref{app_sec:critic_sampling} for step-by-step explanation).
For each problem, we generate $N$ programs, each of which is passed to example unit tests that are often embedded as parts of problem specifications. 
Among the $N$ programs, we denote those that pass example tests as a set $\mathcal{P}$ and the remaining failed programs as a set $\mathcal{F}$.

\paragraph{Program Refining.}
Note that although programs in $\mathcal{P}$ successfully pass example tests, it is not guaranteed that these programs will succeed against the final hidden tests. 
Hidden tests are often more comprehensive and may contain corner cases that challenge these programs. 
Therefore, we can apply another round of generation to further refine the programs.

Specifically, we use sub-sequences from these program samples from $\mathcal{P}$ as prompts (or ``seed'' sequences) to the actor LM. We employ a separate critic model ($\phi_\text{test}$) to guide our choice of sub-sequences from these filtered samples. This critic model is trained with a similar objective as Eq. (\ref{eq:critic_obj}), but in a binary classification setup with $\{\mathrm{FailedTest, PassedTest}\}$ labels.
Let $W^\mathrm{pass} = \{ w_{1}, \dots, w_T \}$ denote a generated sample that passes the example unit tests. We use the critic model to assign a value to each token:
\begin{align}
\hat{q}_{\phi_{\text{test}}}(w_t) = p_{\phi_\text{test}}(\hat{u}=\mathrm{PassedTest} | w_{1:t}, D)
\end{align}
corresponding to the critic's predicted probability of the sub-sequence till $t$ passing the unit tests. 
We split the sequence at position $t_\text{max}$ corresponding to the highest critic assigned value and use the left split as the seed for the next stage. 
If this seed sequence till $t_\text{max}$ contains a token with $p_{\phi_{\text{test}}}(\mathrm{FailedTest})$ $>$ $p_{\phi_{\text{test}}}(\mathrm{PassedTest})$, we further chop it at this token by removing tokens on the right. This is done to pick prompts that are likely to generate successful programs in the next round.

We use these seeds to initialize and condition the (actor) LM to resample new tokens till we encounter the \emph{\textless endoftext \textgreater} token. 
In this round, each seed sequence can be stacked $N/|\mathcal{P}|$ times for upsampling. This results in the same number of output programs $N$ as in the first round of generation.
Finally, we evaluate these $N$ refined programs against the hidden unit tests.

\paragraph{Program Repairing.}
Generating programs to solve a problem, especially a competition-level programming problem, involves a huge search space of possible programs.
Very often, we observe complete failure where all programs fail against example tests, i.e. $|\mathcal{F}|=N$.
Therefore, for these cases, we employ an additional generation step to first repair programs before refining them. 

Specifically, We use the same critic model that is also employed in program refining, to sample top candidates from the set $\mathcal{F}$.
Let $W^\mathrm{fail}$ denote a generated sample that fails the example unit tests. We use the critic model to assign a value to this sample: 
\begin{align}
    \hat{q}_{\phi_{\text{test}}}(W^\mathrm{fail}) = p_{\phi_\text{test}}(\hat{u}=\mathrm{PassedTest} | W^\mathrm{fail}, D)
\end{align}
corresponding to the critic's predicted probability of the the program passing the unit tests. 
We select top $M$ programs with the highest probabilities and pass them to a program repair model $\omega$.

This program repair model is designed as a sequence-to-sequence generation model.
The input sequence is the concatenation of the problem description $D$ and buggy program $W^\mathrm{fail}$. 
We also include additional signals received from the unit test results, include the type of test outcomes (as defined in the return definitions in Eq. (\ref{eq:return1}) to (\ref{eq:return2}), and error subtypes (e.g. syntax errors, out-of-index errors).
The error types are extracted from error traces returned by the compiler.
To train the program repair model, we exploit the synthetic samples that is originally used in our RL training,
as the buggy programs $W^\mathrm{fail}=W^{s}$.
The ground-truth program $W$ can be used as the expected correct program. 
The training objective of the program repair model is to minimize the cross-entropy loss: 
\begin{align}
    \mathcal{L}^\mathrm{repair}_{ce}(\omega) = - \sum_t \log p_\omega(W|D, W^\mathrm{fail}, u, c) = - \sum_t \log [p_\omega (w_t | w_{1:t-1}, D, W^\mathrm{fail}, u, c)]
\end{align}
where $u$ is one of $\{\mathrm{CompileError, RuntimeError, FailedTest, PassedTest}\}$ and $c$ is the error subtype.
During test time, each selected failed sequence can be stacked $N/M$ times for upsampling.
This results in the same number of output programs $N$ as in the first round of generation.
Finally, we pass these $N$ repaired programs and apply the code refining procedure as before. 

\paragraph{Critic Sampling.}
We call the dual strategy of program repairing and refining as ``Critic Sampling'' (CS). 
This dual strategy allows models to generate and improve programs during inference, both from success cases (program refining), and from failure cases (program repairing). 
In practice, we use mini-batch generating to improve efficiency during inference and employ nucleus sampling with a batch size of $N=200$. 
Note that during program refining, while we do incur additional computation costs to re-sample using the seed sequences, we are only required to generate partial programs in the re-generation stage, making this stage less expensive than conventional generation.

\subsection{Implementation Details}
Due to the potential large number of trajectories (i.e. $\mathcal{V}^{T}$) to generate a sequence and the  unstable feedback loop between actor and critic \citep{lillicrap2015continuous, wang2018video}, we applied imitation learning to first warm-start a pretrained LM model with $\mathcal{L}_{ce}$ only for up to $10$ epochs. 
We then sampled program sequences from this actor network to train the critic while keeping the parameters of the actor network frozen. For experiments with CodeT5 actor models, we use the CodeT5-small architecture for the critic model, and GPT2-small critic architecture when the actor models are GPT variants. 
Following \citep{bahdanau2016actor}, since our RL method is applied in a supervised learning task, in addition to synthetic programs, we also use the ground-truth programs of training samples to train the critic.
These samples are considered perfect programs and always have a label of \emph{PassedTest}. 
After training the critic, we then apply both $\mathcal{L}_{ce}$ and $\mathcal{L}_{rl}$ with equal weights to finetune the actor network. 
To optimize the LM actor network, in practice, following previous work \citep{bahdanau2016actor, rennie2017self, wang2018video}, 
in each training optimization step, we can simply approximate the expected gradient with a single sample $W_s \sim p_\theta$:
\begin{align}
    \nabla_\theta \mathcal{L}_{rl}(\theta) \approx - (r(W^s)-r(W^b))  \sum_t \hat{q}_\phi(w^s_t) \nabla_\theta \log p_\theta(w^s_t|w^s_{1:t-1}, D)
\end{align}

\section{Experiments}
\label{sec:experiments}
\subsection{Experimental Setups}

\paragraph{Pretraining.} 
We pretrain a CodeT5-large model (770M) from scratch following T5-large's architecture~\citep{t5}. We follow the pretraining setups in CodeT5~\citep{codet5} with the modifications as proposed in Section \ref{subsec:pretrain_lm}.
Specifically, we adopt the code-specific tokenizer from \citet{codet5}.
We employ 6 programming languages (PLs) in CodeSearchNet~\citep{csn} (CSN) instead of 8 PLs in CodeT5 as C/C\# datasets are not publicly available. 
We apply only the pretraining task of masked span prediction (MSP) from \citep{codet5} and hence, do not have to parse programs into abstract syntax trees (ASTs) to obtain the identifier information.
The last preprocessing step was required in other original pretraining tasks like masked identifier prediction~\citep{codet5}.
To further speed up training, we concatenate data samples to batch size 512 for pretraining with MSP and the resulting number of tokens is 1.1B.
To validate the benefit of using this new pretrained CodeT5 as our foundation model, we evaluate this model on CodeXGLUE \citep{codexglue} and achieve new SOTA results (see Appendix \ref{app_sec:codexglue}). 

We perform our experiments on a kubernetes with 16 A100-40G GPUs on Google Cloud Platform and the total pretraining duration is around $21$ days.
In the first pretraining stage with MSP, we employ a corruption rate of 15\%, a peak learning rate (LR) of 2e-4, and a batch size of $2048$. We pretrain on CSN for $150$ epochs ($10$ days) and then on GCPY for $10$ epochs ($5$ days). 
For the second stage pretraining with NTP, we adopt a peak LR of 1e-4 and a batch size of $256$, and pretrain for $10$ epochs (6 days). We set the maximum length to $768$ and $600$ for source and target sequences respectively for this objective.
For all  experiments, we employ an AdamW optimizer~\citep{DBLP:conf/iclr/LoshchilovH19} with a $0.05$ weight decay and a linear decay LR scheduler with a warmup step of $1000$. 

\paragraph{Evaluation.} 
We follow \citep{hendrycksapps2021, chen2021evaluating} and evaluate the models using the \emph{pass@k} metric, which is the percentage of problems solved by using \emph{k} generated programs per problem.  
We also follow \citep{li2022competition} and use \emph{n@k} metric which only considers a subset of \emph{n} candidates from \emph{k} generated programs per problem. 
The subset of $n$ candidates are typically selected by a filtering method by passing generated programs through \emph{example tests} given as part of the problem description \citep{chen2021evaluating, li2022competition}. 
Note that in our critic sampling procedure, we can perform multiple rounds to refine/repair the programs based on their initial test results. 
In practice, we limit to maximum one round of repairing and/or refining only. 



\subsection{Program Synthesis Benchmarks}

\paragraph{APPS Benchmark.}
We choose the challenging APPS program synthesis benchmark \citep{hendrycksapps2021}, as it has large coding problems of varying difficulties collected from multiple coding websites. 
APPS consists of 10,000 coding problems with a 50-50 train-test split. Each problem is accompanied by 23.2 correct Python programs and 21.2 unit tests on average. The average length per problem is 293.2 words and the average length per program is 18.0 lines. The dataset is categorized into three levels of difficulty: Introductory (3639, train/test=2639/1000), Interview (5000, train/test=2000/3000), and Competition (1361, train/test=361/1000). 
Each sample includes $20$ unit tests on average to validate the functional correctness of programs. 
We follow the same preprocessing step as \citep{hendrycksapps2021} to formulate the input sequences from problem descriptions.

On APPS, we finetune our pretrained CodeT5 following our CodeRL framework (Section \ref{subsec:apps}).
To warm-start CodeT5 models with $\mathcal{L}_{ce}$, we adopt a batch size of $64$ and warmup LR from 0 to 2e-5 for the first 500 steps and polynomially (power=$0.5$) decay to 1e-5 until the end of $10$ epochs, which takes around $30$ hours on one A100 GPU. We set the maximum source and target sequence length to $600$ and $512$ respectively.

\paragraph{MBPP Benchmark.}
We additionally include another smaller and simpler Python program synthesis dataset called MBPP~\citep{austin2021program} (Mostly Basic Programming Problems) for evaluation. The dataset contains 974 instances with 374/90/500 instances for training/validation/testing respectively and 10 reserved for few-shot learning. 
The problems are typically short, usually one sentence of natural language descriptions each.
Each problem is accompanied by 1 correct solution (6.8 lines of code on average) and 3 unit tests in the form of \texttt{assert} statements for validating the functional correctness. Unlike APPS, unit tests in MBPP are not hidden and are explicitly incorporated into the source sequences for program synthesis models. 
This might encourage models to be overfitting to these \texttt{assert} statements via hard-coding an if-expression very occasionally.
However, for a fair comparison with the baselines, we construct the source sequences in the same way as prior work. Specifically, we adopt the same prompt format as~\citep{austin2021program} to prepare the input sequence as: problem descriptions + ``Your code should satisfy these tests:'' + 3 \texttt{assert} statements.

On MBPP, we experiment with in both zero-shot (Section \ref{subsec:mbpp}) and full finetuning (Appendix \ref{app_sec:mbpp_results}) setup. 
To finetune CodeT5, due to the small training set of MBPP, we finetune the models for $60$ epochs with a constant LR of 2e-5 and a batch size of $32$, which takes less than $30$ mins on one A100. We set the maximum source and target  length to $382$ and $306$ respectively.

\subsection{Experimental Results on APPS}
\label{subsec:apps}
\paragraph{Baselines.} 
As reported by \citet{hendrycksapps2021}, we compared our models with several baselines, including GPT2 \citep{radford2019language}, GPT-Neo \citep{black10gpt}, and GPT3 \citep{brown2020language}.
We also compare the results with Codex \citep{chen2021evaluating} and AlphaCode \citep{li2022competition}.
Note that by default, results of pretrained LMs (except for Codex and GPT3) are from models finetuned on APPS using the standard loss $\mathcal{L}_{ce}$ only.
In our ablations, since CodeRL is model-agnostic, we can also integrate it with GPT variants such as GPT-J \citep{gpt-j} and GPT-Neo.

\begin{table}[t]
\centering
\caption{\textbf{Results on the APPS benchmark}: 
Overall, CodeRL can bring the performance gains of CodeT5 models and achieves new SOTA on both \emph{pass@k} and \emph{n@k} metrics.
``Intro'': introductory, ``Inter'': interview, ``Comp'': competition-level tasks.
}
\begin{subtable}[htbp]{1.0\textwidth}
\centering
\caption{Performance by \emph{pass@k} with $k=\{1,5,1000\}$}
\label{tab:pass_k}
\resizebox{1.0\textwidth}{!} {
\begin{tabular}{lc|cccc|cccc|cccc}
\hline
\multicolumn{1}{c}{}                        &                        & \multicolumn{4}{c|}{\emph{pass@1}}                                       & \multicolumn{4}{c|}{\emph{pass@5}}                                        & \multicolumn{4}{c}{\emph{pass@1000}}                                        \\ 
\cline{3-14}
\multicolumn{1}{c}{\multirow{-2}{*}{Model}} & \multirow{-2}{*}{Size} & Intro         & Inter         & Comp          & All           & Intro          & Inter         & Comp          & All           & Intro          & Inter          & Comp           & All            \\\hline
Codex                                       & 12B                    & 4.14          & 0.14          & 0.02          & 0.92          & 9.65           & 0.51          & 0.09          & 2.25          & 25.02          & 3.70           & 3.23           & 7.87           \\
AlphaCode                                   & 1B                     & -             & -             & -             & -             & -              & -             & -             & -             & 17.67          & 5.24           & 7.06           & 8.09           \\
GPT3                                        & 175B                   & 0.20          & 0.03          & 0.00          & 0.06          & -           & -          & -          & -          & -              & -              & -              & -              \\

GPT2                                        & 0.1B                   & 1.00          & 0.33          & 0.00          & 0.40          & 2.70           & 0.73          & 0.00          & 1.02          & -              & -              & -              & -              \\
GPT2                                        & 1.5B                   & 1.30          & 0.70          & 0.00          & 0.68          & 3.60           & 1.03          & 0.00          & 1.34          & 25.00              & 9.27              & 8.80              & 12.32              \\
GPT-Neo                                     & 2.7B                   & 3.90          & 0.57          & 0.00          & 1.12          & 5.50           & 0.80          & 0.00          & 1.58          & 27.90              & 9.83              & 11.40              & 13.76              \\
GPT-J                                       & 6B                     & 5.60          & 1.00          & 0.50          & 1.82          & 9.20           & 1.73          & 1.00          & 3.08          & 35.20              & 13.15              & 13.51              & 17.63              \\
 \hline
CodeRL+CodeT5                                 & 770M                   & \textbf{7.08}	& \textbf{1.86} & 	\textbf{0.75} &	\textbf{2.69} &	\textbf{16.37}	& \textbf{4.95}	& \textbf{2.84}	& \textbf{6.81} &	\textbf{40.00} &	\textbf{15.67}	& \textbf{17.90}	& \textbf{20.98} \\

\hline
\end{tabular}
}
\end{subtable}

\begin{subtable}[htbp]{1.0\textwidth}
\centering
\caption{Performance by \emph{n@k} with $k$ up to 50000 and $n=\{1,5\}$}
\label{tab:n_pass_k}
\resizebox{1.0\textwidth}{!} {
\begin{tabular}{lcc|cccc|cccc}
\hline
\multicolumn{1}{c}{}                        &                        &                     & \multicolumn{4}{c|}{\emph{1@k}}                                        & \multicolumn{4}{c}{\emph{5@k}}                                         \\
\cline{4-11}
\multicolumn{1}{c}{\multirow{-2}{*}{Model}} & \multirow{-2}{*}{Size} & \multirow{-2}{*}{$k$} & Intro          & Inter         & Comp          & All           & Intro          & Inter         & Comp          & All            \\
\hline
Codex                                       & 12B                    & 1000                & \textbf{22.78} & 2.64          & 3.04          & 6.75          & 24.52          & 3.23          & 3.08          & 7.46           \\
AlphaCode                                   & 1B                     & 1000                & -              & -             & -             & -             & 14.36          & 5.63          & 4.58          & 7.17           \\
AlphaCode                                   & 1B                     & 10000               & -              & -             & -             & -             & 18.18          & 8.21          & 6.65          & 9.89           \\
AlphaCode                                   & 1B                     & 50000               & -              & -             & -             & -             & 20.36          & \textbf{9.66} & 7.75          & 11.42          \\
\hline

CodeRL+CodeT5                                 & 770M                   & 1000                & 17.17 &	\textbf{6.78} &	 \textbf{4.88}	 & \textbf{8.48}	 & \textbf{25.61}	 &9.53	 &\textbf{8.91}	 & \textbf{12.62}   \\

\hline
\end{tabular}
}
\end{subtable}
\end{table}

\paragraph{Overall Results.}
Firstly, Table \ref{tab:pass_k} shows that the CodeRL with the CodeT5 model can achieve significant performance gains, outperforming many pretrained LMs of much larger sizes. 
Specifically, our approach achieved new SOTA results of $2.69$\% \emph{pass@1}, $6.81$\% \emph{pass@5}, and $20.98$\% \emph{pass@1000}.
Table \ref{tab:n_pass_k} shows that 
when evaluating on a subset of filtered code samples, our CodeRL+CodeT5 can achieve SOTA results of $8.48$\% \emph{1@k} and $12.62$\% \emph{5@k}. 

Secondly, similar to prior work \cite{hendrycksapps2021, austin2021program, chen2021evaluating}, we also observe the benefits of upsampling generation when increasing the number of generation samples $k$ from $1$ to $1000$.  
Note that while CodeRL incurs additional computation cost during inference with CS, our approach only requires much lower $k$ to achieve comparable performance with other models. 
Specifically, with $k=1000$ only, our model performance is as good as AlphaCode with much a larger generation budget of $k=50000$.
Finally, from Table \ref{tab:n_pass_k}, we found that for challenging programming tasks in interview and competition levels, finetuning can significantly improve model performance. 
Specifically, we note that Codex, which was not finetuned on APPS and tested in a few-shot setting, can achieve good \emph{n@1000} results, but the model fails dramatically at synthesis tasks in interview and competition levels.
This observation indicates a significant gap between the pretraining stage and downstream synthesis tasks. 

\subsection{Ablation Studies}

In this section, for a fair comparison between variants of return estimates and learning objectives, we report the results of \emph{pass@k} where $k=\{1,5\}$ with beam search decoding.
For ablation analysis of CodeRL during inference with larger $k$, we report the results with and without the CS procedure.

\paragraph{Impacts of Return Estimates.}
Table \ref{tab:codet5_ablation_return} show the results of CodeT5-770M trained by different approaches to estimate returns of code samples.
Overall, we report that the CodeRL objective with relative token-level return estimates by our critic model (Model D) can achieve the best performance on \emph{pass@1} and \emph{pass@5}.
Secondly, we note that using absolute returns without a baseline (Model B) could lead to the most performance drop, as this approach heavily penalizes all incorrect samples (even though they might still be better than a naive baseline).
Hence, considering relative return estimates that can effectively exploit imperfect codes can lead to better synthesis systems. 

Thirdly, without a critic model, simply assigning identical rewards to all tokens in a code sample (Model A) is disadvantageous as these return estimates are too restrictive to be used as feedback signals for RL training. 
For instance, a program is considered incorrect only because of an additional blank space character, which can result in a \emph{Indentation Error} in a Python program. 
Simply assigning an identical reward to all tokens in this program will heavily penalize correct parts of the program sequence. 
Finally, we experimented with a distance-based critic which assumes that token values $\hat{q}(w^s_t)$ decay linearly from $t=1$ to $t=T$ (Model C). 
The lower performance suggests the benefit of training a critic network to compute the returns rather than relying on rule-based approaches. 

\begin{table}[t]
\centering
\caption{
\textbf{Ablation results with variants of return estimates:}
CodeT5 model that is trained with return estimates using a baseline ($W_b$) and a trained critic model $\hat{q}_\theta$ can achieve the best performance. 
``dist.'' indicates a rule-based approach that estimates returns following a linear decay by token positions from $t=1$ to $t=T$. 
}
\label{tab:codet5_ablation_return}
\begin{tabular}{ccc|cccc|cccc}
\hline
\multirow{2}{*}{\#}  & \multirow{2}{*}{$W^b$} & \multirow{2}{*}{$\hat{q}_\phi$} & \multicolumn{4}{c|}{\emph{pass@1}}     & \multicolumn{4}{c}{\emph{pass@5}}     \\
\cline{4-11}
&                                                                        &                                 & Intro & Inter & Comp & All  & Intro & Inter & Comp & All  \\
                                                                        \hline
A                                                                 & \checkmark                         & -                       & 4.60  & 1.10  & 0.20 & 1.62 & 7.10  & 1.57  & 0.40 & 2.44 \\
B                                                                    & -                         & \checkmark                       & 4.00  & 0.87  & 0.20 & 1.36 & 5.60  & 1.30  & 0.20 & 1.94 \\
C                                                                    & \checkmark                         & dist.                & 4.90  & 1.03  & 0.20 & 1.64 & 7.80  & 1.60  & 0.30 & 2.58 \\
D                                                                    & \checkmark                         & \checkmark                       & \textbf{6.20}  & \textbf{1.50}  & \textbf{0.30} & \textbf{2.20} & \textbf{9.39}  & \textbf{1.90}  & \textbf{0.42} & \textbf{3.10} \\
\hline
\end{tabular}
\end{table}

\begin{table}[t]

\centering
\caption{
\textbf{Ablation results with different learning objectives:}
We experiment with both CodeT5 and GPT-Neo with different combinations of cross-entropy loss $\mathcal{L}_{ce}$ and reinforcement learning loss $\mathcal{L}_{rl}$. 
Note that these losses are applied on models that are already warm-started with conventional cross-entropy losses for up to 10 epochs. 
} 
\label{tab:codet5_ablation_loss}
\begin{tabular}{cc|cccc|cccc}
\hline
\multirow{2}{*}{$\mathcal{L}_{ce}$} & \multirow{2}{*}{$\mathcal{L}_{rl}$} & \multicolumn{4}{c|}{\emph{pass@1}}                                       & \multicolumn{4}{c}{\emph{pass@5}}                                       \\
\cline{3-10}
                         &                          & Intro         & Inter         & Comp          & All           & Intro         & Inter         & Comp          & All           \\
                         \hline
\multicolumn{10}{c}{\cellcolor[HTML]{EFEFEF}GPT-Neo}                                                                                                                                                                                                                            \\ \hline
-                & -                         & 3.90                      & 0.57                      & 0.00                     & 1.12                    & 5.50                      & 0.80                      & 0.00                     & 1.58                    \\
\checkmark                         & -                         & 2.70                      & \textbf{0.90}             & 0.10                     & 1.10                    & 5.00                      & 1.43                      & 0.30                     & 1.92                    \\
\checkmark(+$W^s$)                    & -                         & 2.90                      & 0.80                      & \textbf{0.30}            & 1.12                    & 5.20                      & \textbf{1.57}             & \textbf{0.40}            & 2.06                    \\
-                         & \checkmark                         & 3.30                      & 0.80                      & 0.20                     & 1.18                    & 5.30                      & \textbf{1.57}             & 0.20                     & 2.04                    \\
\checkmark                         & \checkmark                         & \textbf{4.70}             & 0.73                      & \textbf{0.30}            & \textbf{1.44}           & \textbf{6.58}             & 1.54                      & 0.18                     & \textbf{2.28}          \\
\hline 
\multicolumn{10}{c}{\cellcolor[HTML]{EFEFEF}CodeT5-770M}                                                                                                                                                                                                                            \\ \hline 
-               & -                        & \textbf{6.60} & 1.03          & 0.30          & 2.00          & 8.80          & 1.67          & \textbf{0.70} & 2.90          \\
\checkmark                        & -                        & 4.60          & 0.93          & 0.10          & 1.50          & 7.00          & 1.37          & 0.20          & 2.26          \\
\checkmark(+$W^s$)                   & -                        & 5.10          & 1.10          & 0.40          & 1.76          & 8.30          & 1.43          & \textbf{0.70} & 2.66          \\
-                        & \checkmark                        & 5.00          & 0.90          & \textbf{0.50} & 1.64          & 7.60          & 1.53          & 0.60          & 2.56          \\
\checkmark                        & \checkmark                        & 6.20          & \textbf{1.50} & 0.30          & \textbf{2.20} & \textbf{9.39} & \textbf{1.90} & 0.42          & \textbf{3.10} \\
\hline
\end{tabular}
\end{table}

\begin{figure}[t]
	\centering
	\resizebox{1.0\textwidth}{!} {
	\includegraphics{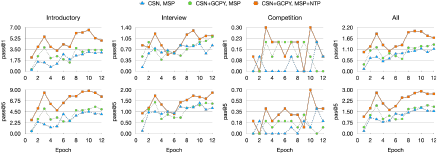}
	}
	\caption{
	\textbf{Ablation results by finetuning epochs}:
	We report the finetuning progress of CodeT5-$770$M models that are pretrained on different configurations by pretraining data and pretraining tasks. CSN: CodeSearchNet, GCPY: Github Code Python, MSP: Masked Span Prediction, NTP: Next Token Prediction. 
	All models are finetuned only with $\mathcal{L}_{ce}$ on APPS.
	}
	\label{app_fig:codet5_apps}
\end{figure}

\paragraph{Impacts of Learning Objectives.} 
Table \ref{tab:codet5_ablation_loss} shows the results with different combinations of $\mathcal{L}_{ce}$ and $\mathcal{L}_{rl}$.
Since CodeRL is model-agnostic, we apply the ablation experiments to both CodeT5 and GPT-Neo \citep{black10gpt}. 
Note that in these experiments,  $\mathcal{L}_{ce}$ and $\mathcal{L}_{rl}$ are applied on models that are already warm-started/finetuned with $L_{ce}$ for up to 10 epochs. 
Firstly, when we experiment with using only $\mathcal{L}_{rl}$, we note the problem of vanishing gradients during finetuning, which was similarly observed by \citet{DBLP:journals/corr/RanzatoCAZ15, bahdanau2016actor}.
Therefore, the final models actually deteriorate and lead to performance drops. 
Secondly, we note that by using only $\mathcal{L}_{ce}$ for further finetuning, despite improvement in losses during training time, the model performance indeed degrades during test time.
We expect these models are overfitting to the training data, as similarly observed in our analysis of pretrained models in Figure \ref{app_fig:codet5_apps}.

Interestingly, we found that a naive approach of $\mathcal{L}_{ce}$ with synthetic samples $W^s$, all of which are treated as correct codes with $r(W^s)=1$, still leads to some performance improvement with GPT-Neo on \emph{pass@5}.
However, in all other cases, this training strategy does not work as well as considering a critic model to estimate returns of $W_s$ by their test results. 
Finally, we found that using both $\mathcal{L}_{ce}$ and $\mathcal{L}_{rl}$ results in a more consistent performance improvement overall on \emph{pass@1} and \emph{pass@5} for the GPT-Neo and CodeT5 models. 


\begin{table}[t]
\centering
\caption{
\textbf{Ablation results of Critic Sampling (CS):}
We experiment with CodeT5 with different combinations of program refining and repairing steps. 
Overall, compared to results without CS, combining both approaches lead to the best program improvement. 
$M$: the number of top candidates selected from program samples that fail example unit tests. 
}
\label{tab:critic_sampling}
\resizebox{1.0\textwidth}{!} {
\begin{tabular}{cc|cccc|cccc|cccc}
\hline
\multicolumn{2}{c|}{Critic Sampling} & \multicolumn{4}{c|}{\emph{pass@200}}  & \multicolumn{4}{c|}{\emph{pass@1000}} & \multicolumn{4}{c}{\emph{1@1000}}  \\
\hline
Refine           & Repair           & Intro & Inter & Comp  & All   & Intro & Inter & Comp  & All   & Intro & Inter & Comp & All  \\
\hline
-                & -                & 26.79 & 8.73  & 7.60  & 12.12 & 35.30 & 13.33 & 13.60 & 17.78 & 16.27 & 6.00  & 4.27 & 7.71 \\
\checkmark                & -                & 29.10 & 9.67  & 9.50  & 13.52 & 38.10 & 14.33 & 15.70 & 19.36 & 16.52 & 6.16  & 4.15 & 7.83 \\
\checkmark                & \checkmark ($M=1$)          & 29.80 & 10.43 & 10.80 & 14.38 & \textbf{40.00} & \textbf{15.67} & 17.90 & \textbf{20.98} & \textbf{17.17} & 6.78  & 4.88 & \textbf{8.48} \\
\checkmark                & \checkmark ($M=2$)          & \textbf{30.20} & 10.20 & \textbf{11.50} & \textbf{14.46} & 39.90 & 15.57 & \textbf{18.00} & 20.92 & 16.96 & \textbf{6.82}  & \textbf{4.90} & 8.47 \\
\checkmark                & \checkmark ($M=4$)          & 29.50 & \textbf{10.60} & 10.80 & 14.42 & 39.40 & 15.37 & 17.60 & 20.62 & 16.99 & 6.63  & 4.78 & 8.33 \\
\hline
\end{tabular}
}
\end{table}

\paragraph{Impact of Critic Sampling.}
Table \ref{tab:critic_sampling} shows the ablation results of critical sampling (CS) during inference, applied on CodeT5 models. 
We experiment with different combinations of program refining and repairing steps.
Overall, we found positive impacts of CS, combining both program refining and repairing, across all metrics, with particularly more significant gains on \emph{pass@1000}. 
We note that just program refining alone can help to bring performance gains, but its impact is reduced on the $1@1000$ metric. 
Note that $n@k$ measures the solving rate among the subset $\mathcal{P}$ filtered from $k$ samples. 
As program refining will technically increase the size of this subset, the \emph{n@k} metric will consider an exponentially larger number of options of $n$ samples than before. 
This will normalize $n@k$ by a larger pool of $n$ candidate set, resulting in less impact of program refining on model performance.
We recommend additional post-processing steps such as candidate ranking \citep{cobbe2021training} to improve the performance of program refining, particularly on \emph{n@k} metrics. 

Secondly, when integrated program refining with program repairing (for problems where $\mathcal{P}=\emptyset$), we found further performance gains in all metrics. 
Interestingly, when experimenting with different top-$M$ selection schemes, we found the best overall performance with $M=1$ and performance starts to drop from $M=2$ to $M=4$ (except for \emph{pass@200} results). 
This observation indicates the benefit of using the critic model to focus on the best candidates for program repairing rather than choosing multiple program candidates.  
Moreover, with larger $M$, each program candidate will have a smaller number of batch size (i.e. $N/M$).
This results in a lower chance for the program repair model to properly repair and generate correct programs. 

\paragraph{Results on Competition-level Tasks.}


\begin{figure}[t]
	\centering
	\includegraphics[width=1\textwidth]{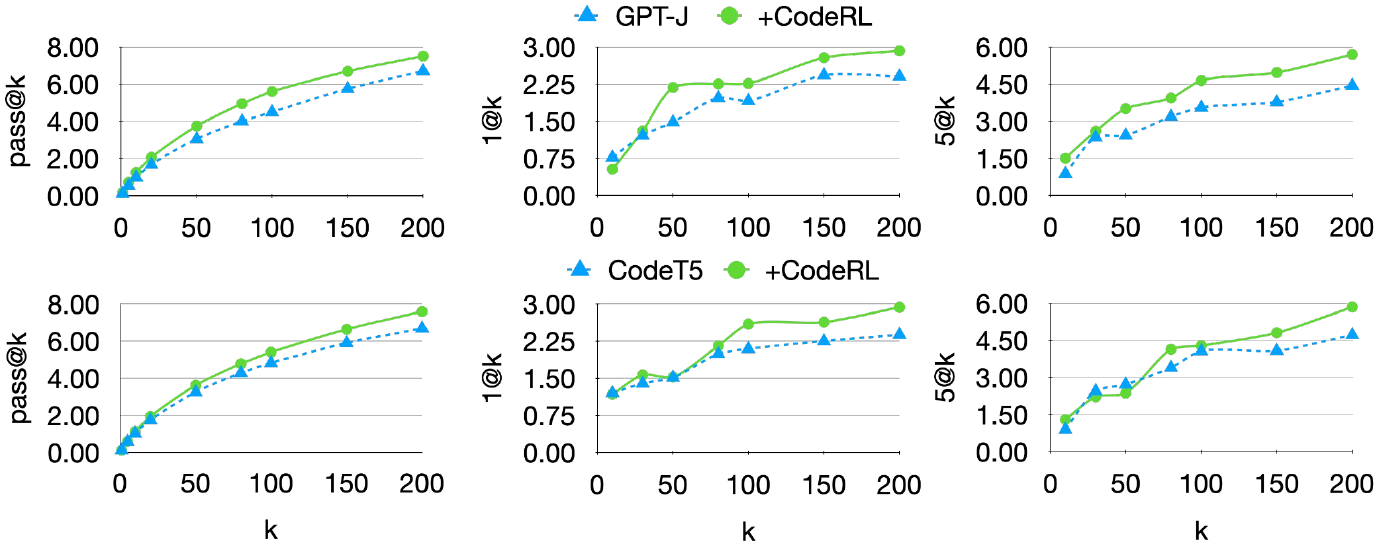}
  \caption{
  \textbf{Results on APPS competition-level test samples:}
  We investigate the most challenging programming problem tasks, i.e. competition level, in the APPS benchmark. 
  Integrating CodeRL with both CodeT5 and GPT-J, we observe good performance improvement across $pass@k$ and $n@k$ metrics, with increasing performance gains as $k$ increases. 
  }
  \label{fig:pass_k_competition}
\end{figure}

We choose to investigate a subset of the APPS test split, which contains the test samples of the highest difficulty level (i.e. competition programming tasks). 
Figure \ref{fig:pass_k_competition} shows the results of \emph{pass@k} and \emph{n@k} with \emph{k} ranging from $1$ to $200$ and $n=\{1,5\}$, for CodeRL+CodeT5 and CodeT5 only.
Since CodeRL is model-agnostic, we also integrate it with GPT-J \citep{gpt-j} and report the results. 
To focus on the impact of our RL optimization, during test time, we compare models using only nucleus sampling and without the CS procedure. 
Figure \ref{fig:pass_k_competition} shows that the performance gains are quite consistent on both GPT-J and CodeT5. 
In particular, as $k$ increases, the performance gain of CodeRL is more significant on both GPT-J and CodeT5 models.
We attribute these gains to the CodeRL learning objective $\mathcal{L}_{rl}$ that encourages models to explore code solutions drawn from the model's sampling distribution. 
During test time with an increasing $k$ sampling budget, models are allowed to generate diverse code solutions and the impact of $\mathcal{L}_{rl}$ becomes more significant.

\begin{table}[t]  
\centering
\caption{
\textbf{Ablation results of CodeT5 pretrained model variants:}
We report the results of models pretrained on different configurations by model size, pretraining data, and pretraining task. 
CSN: CodeSearchNet, GCPY: Github Code Python, MSP: Masked Span Predition, NTP: Next Token Prediction. 
For a fair comparison, all models are finetuned only with $\mathcal{L}_{ce}$ on APPS. 
}
\label{tab:codet5_ablation}
\begin{tabular}{ccc|cccc|cccc}
\hline
\multirow{2}{*}{Size} & \multirow{2}{*}{\begin{tabular}[c]{@{}c@{}}Data\\\end{tabular}} & \multirow{2}{*}{\begin{tabular}[c]{@{}c@{}}Task\\\end{tabular}} & \multicolumn{4}{c|}{\emph{pass@1}}     & \multicolumn{4}{c}{\emph{pass@5}}     \\
\cline{4-11}
                      &                                                                                  &                                                                             & Intro & Inter & Comp & All  & Intro & Inter & Comp & All  \\
                      \hline
60M                   & CSN                                                                              & MSP                                                                         & 1.40  & 0.67  & 0.00 & 0.68 & 2.60  & 0.87  & 0.10 & 1.06 \\
220M                  & CSN                                                                              & MSP                                                                         & 2.50  & 0.73  & 0.00 & 0.94 & 3.30  & 1.10  & 0.10 & 1.34 \\
770M                  & CSN                                                                              & MSP                                                                         & 3.60  & 0.90  & 0.20 & 1.30 & 4.30  & 1.37  & 0.20 & 1.72 \\
770M                  & +GCPY                                                                            & MSP                                                                         & 4.30          & \textbf{1.10} & 0.20          & 1.56          & 5.60          & 1.47          & 0.30          & 2.06          \\
770M                  & +GCPY                                                                            & +NTP                                                                        & \textbf{6.60} & 1.03          & \textbf{0.30} & \textbf{2.00} & \textbf{8.80} & \textbf{1.67} & \textbf{0.70} & \textbf{2.90} \\
\hline
\end{tabular}
\end{table}

\paragraph{Impacts of Pretraining Approaches for CodeT5.}
As commonly observed in prior work \citep{austin2021program, chen2021evaluating, li2022competition}, the performance of synthesis systems is correlated with the quality of foundation models. 
In Table \ref{tab:codet5_ablation}, we report the results of CodeT5 with different configurations of model sizes, pretraining data, and pretraining objectives. 
For a fair comparison, all models are only finetuned/ warm-started with $\mathcal{L}_{ce}$ on APPS up to 12 epochs. 
As similarly observed in prior work \citep{chen2021evaluating, austin2021program}, we found that scaling up the number of model parameters (from $60$M to $770$M) can significantly improve model performance of downstream synthesis tasks. 
When we improve the pretraining data by adding the GCPY dataset (10x larger than the CSN dataset), we also observe good performance improvement, i.e. from $1.3$ to $1.56$ \emph{pass@1}, and $1.72$ to $2.06$ \emph{pass@5}. 
Finally, by combining the pretraining objective from Masked Span Prediction (MSP) and Next Token Prediction (NTP), the model is able to adapt better to the downstream synthesis task.
Notably, adding NTP pretraining task can improve the performance from $2.06$ to $2.9$ \emph{pass@5}. 

\paragraph{Results by Finetuning Epochs with NTP Objective.}
Figure \ref{app_fig:codet5_apps} shows the performance of CodeT5 model variants by finetuning epochs and by difficulty levels of programming tasks. 
Note that in these experiments, we only need to compare among CodeT5 model variants by pretraining strategies, and hence, only engage $\mathcal{L}_{ce}$  in the finetuning stage on APPS. 
Consistent with our prior analysis, enhancing both pretraining data (with larger data of GCPY) and pretraining objectives (with NTP objective) improves model performance across training epochs in general. 
Moreover, as noted by our analysis of learning objectives, only using $\mathcal{L}_{ce}$ often leads to overfitting performance, typically after epoch $10$ in our case.
Hence, to further finetune large-scale LMs, we recommend adopting our RL objective $\mathcal{L}_{rl}$ to utilize synthetic training samples and avoid overfitting models.

\begin{table}[t] 
\centering
\caption{
\textbf{Results on the MBPP benchmark:}
we test the zero-shot transfer ability of CodeRL. 
CodeRL+CodeT5 (ZS) which was trained on APPS with $\mathcal{L}_{rl}$ and evaluated on MBPP (Mostly Basic Programming Problems) Benchmark \citep{austin2021program} in a zero-shot setting, achieves new SOTA. 
}
\label{tab:codet5_mpbb}
\begin{tabular}{lc|c}
\hline
Model & Size & \emph{pass@80} \\
\hline
GPT & 224M & 7.2 \\
GPT & 422M & 12.6 \\
GPT & 1B & 22.4 \\
GPT & 4B & 33.0 \\
GPT & 8B & 40.6 \\
GPT & 68B & 53.6 \\
GPT & 137B & 61.4 \\
\hline
CodeRL+CodeT5 (ZS) & 770M & \textbf{63.0} \\
\hline
\end{tabular}
\end{table}

\begin{figure}[t]
	\centering
	\resizebox{1.0\textwidth}{!} {
	\includegraphics{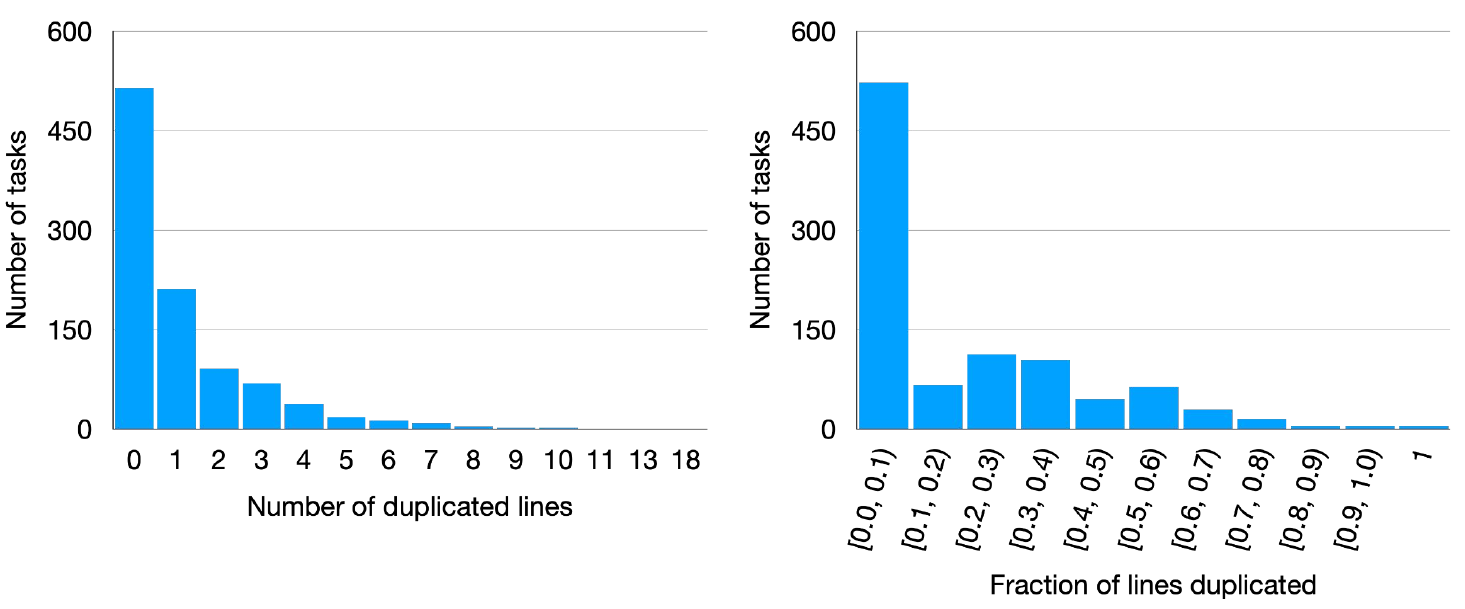}
	}
	\caption{
	\textbf{Analysis of duplicated lines between MBPP and APPS:}
	The overlap of data between APPS and MBPP appears to be minimal, with only $12.6\%$ MBPP programs with $>50\%$ lines duplicated in APPS training data. 
	}
	\label{fig:duplicated_lines}
\end{figure}

\subsection{Experimental Results on MBPP}
\label{subsec:mbpp}

\paragraph{Zero-shot evaluation on MBPP Benchmark.}
In addition to the APPS benchmark, we test the zero-shot transfer ability of CodeRL on another smaller and simpler program synthesis benchmark MBPP~\citep{austin2021program}. Table~\ref{tab:codet5_mpbb} reports the results of our CodeRL+CodeT5 on MBPP benchmark compared with finetuned GPT models of up to 137B size. Our CodeRL+CodeT5 (ZS) was trained on APPS and then evaluated on MBPP in a zero-shot setting. We observe that CodeRL with CodeT5 of a much smaller model size yields surprisingly good zero-shot performance, setting a new SOTA result of 63.0\% \emph{pass@80} over GPT-137B's 61.4\% \emph{pass@80}. This validates the strong zero-shot transfer ability of CodeRL for unseen tasks. 

\paragraph{Overlap between MBPP and APPS.}
A common concern about transfer learning is that the source (APPS) and target (MBPP) tasks might have overlap in their training data, which could result in the source model tending to memorize these substantially similar data when applied to the target task. To address this concern, we analyze how many lines of code appear in both the training set of APPS and programs of MBPP following~\citet{austin2021program}. For this analysis, we discard code comments and normalize the whitespaces for each line,  and then exclude lines that appear more than twice anywhere in MBPP, as these are likely to be common Python keywords such as \texttt{return} and \texttt{break}.

Figure~\ref{fig:duplicated_lines} illustrates the number of absolute duplicated lines (Left) and relative fraction of duplicated lines (Right) in the MBPP programs. As can be seen, the overlap between APPS and MBPP seems to be minimal. Only 12.6\% MBPP programs have more than half of their lines matched somewhere in the APPS training data.
Besides, more than half (514 out of 974) of programs have a zero overlap and 90.9\% have only no more than 3 lines overlapped with the APPS training set. If we further require the  lines to be consecutive, there are no more than 2 consecutive duplicated lines. 
More experimental analysis is included in Appendix \ref{app_sec:mbpp_results}. 


\begin{figure}[t]
     \centering
     \begin{subfigure}[t]{1.0\textwidth}
         \centering
         \includegraphics[width=\textwidth]{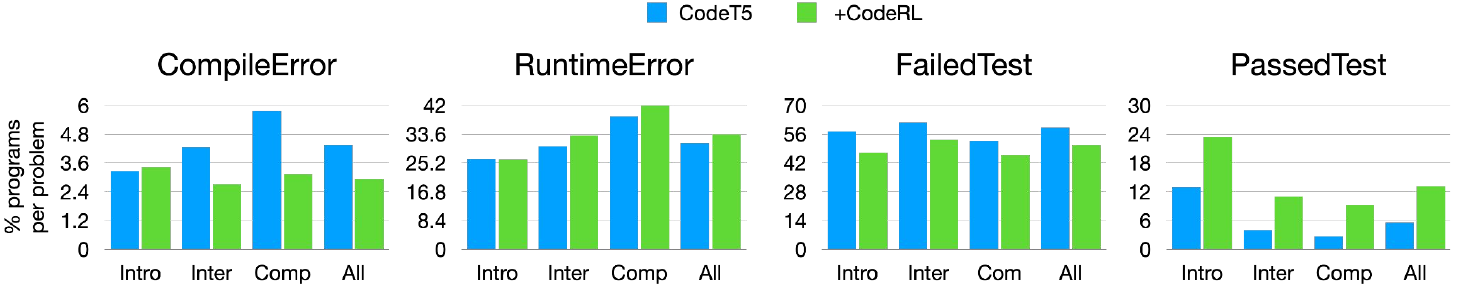}
         \caption{Results on example unit tests}
        \label{app_fig:qual_result_example}
     \end{subfigure}
     \begin{subfigure}[t]{1.0\textwidth}
         \centering
         \includegraphics[width=\textwidth]{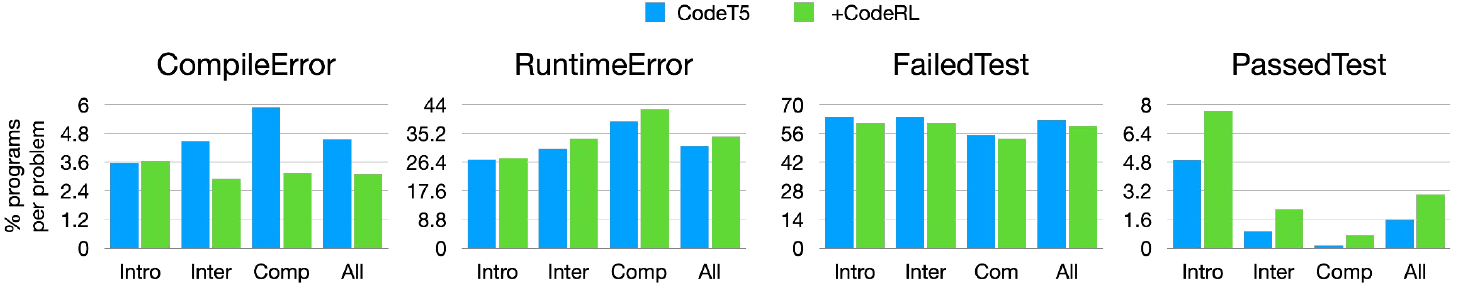}
         \caption{Results on hidden unit tests}
         \label{app_fig:qual_result_hidden}
     \end{subfigure}
        \caption{
        \textbf{Qualitative results by test outcomes: }
	Using CodeT5 models with and without CodeRL, we generate 200 programs per test sample on APPS and report the \% programs per sample by their unit test outcomes, including CompileError, RuntimeError, FailedTest, and PassedTest. 
	Test outcomes are defined accordingly to our definition in Eq. (\ref{eq:return1}) to (\ref{eq:return2}).
	}
        \label{app_fig:qual_result}
\end{figure}

\subsection{Qualitative Analysis}

\paragraph{Analysis by Unit Test Outcomes.}
Figure \ref{app_fig:qual_result} demonstrates the average percentages of generated programs per problem, grouped by their test outcomes.
Specifically, we use CodeT5 or CodeRL+CodeT5 to generate programs and randomly select $200$ generated programs per test sample in the APPS test split.
We pass programs to either example unit tests or hidden unit tests and group the output programs by their test outcomes.
The outcomes are categorized according to our definition in Eq. (\ref{eq:return1}) to (\ref{eq:return2}), including \emph{CompileError, RuntimeError, FailedTest, and PassedTest}. 

First, on both example unit tests and hidden unit tests, we observe that integrating CodeRL can increase the likelihood that a program can pass the tests, and reduces the probability that it fails one or more unit tests.
The probability to pass unit tests are improved more significantly in introductory-level programming problems. 

Secondly, we note that the percentages of having compiling errors decrease in CodeRL-generated programs, with more effects on interview and competition-level problems. 
As compiling errors are less likely to occur with CodeRL programs, these programs are still suffered from runtime errors.
This leads to a higher probability that a CodeRL program contains runtime errors. 
More analysis of compiling and runtime failures is described in Appendix \ref{app_sec:failure_analysis}. 

We note that there are quite significant performance gaps by test outcomes between example unit tests (Figure \ref{app_fig:qual_result_example}) and hidden unit tests (Figure \ref{app_fig:qual_result_hidden}).
This observation suggests that example tests are not as comprehensive as hidden tests and hence, limit the positive impacts of our CodeRL generation procedure due to false positives. 
We recommend additional methods to improve example unit tests, such as through data augmentation by mutating input/output pairs \citep{austin2021program}.

\begin{figure}[t]
	\centering
	\resizebox{1.0\textwidth}{!} {
	\includegraphics{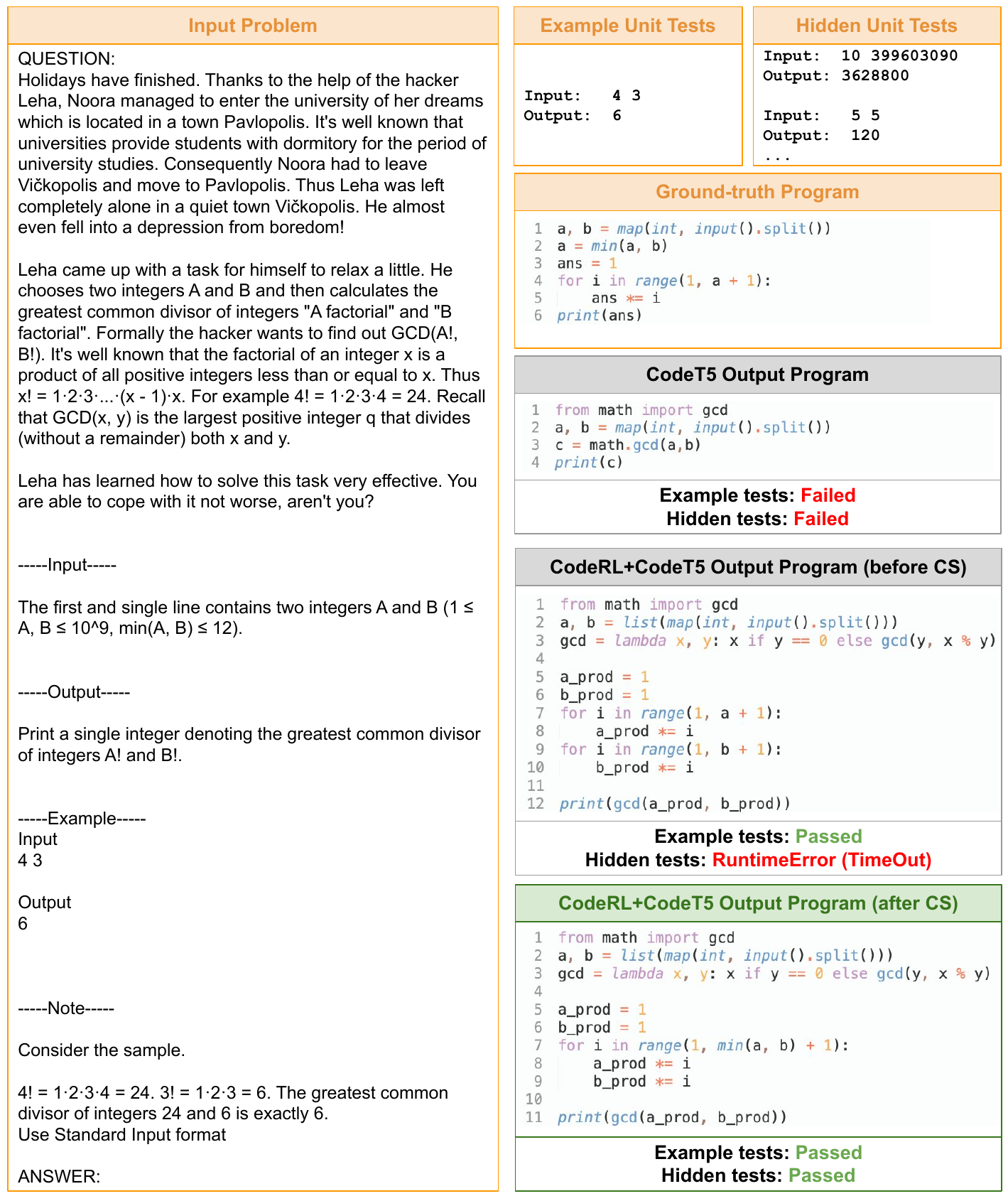}
	}
	\caption{
	\textbf{An example interview-level APPS programming task and programs generated by CodeT5 variants:}
	The program generated by CodeT5 model fails all unit tests while CodeRL+CodeT5 (without CS generation) can generate a functionally correct program.
	However, this program leads to runtime errors in extreme test cases.
	After applying the CS generation procedure, the program is improved and able to pass all hidden unit tests. 
	}
	\label{app_fig:example_program2}
\end{figure}

\paragraph{Example Generated Program.}
Figure \ref{app_fig:example_program2} shows an example of a programming problem from the APPS benchmark and corresponding programs generated by CodeT5 variants. 
Specifically, based on the same foundation pretrained CodeT5 (pretrained with GCPY data and NTP objective), we compare the CodeT5 model that is finetuned by $\mathcal{L}_{ce}$ only and another model that follows our CodeRL framework.
In CodeRL+CodeT5, we further show programs before and after applying the CS procedure. 
We found that applying CodeRL can generate more appropriate programs and using the CS procedure further improves their functional correctness. 
For instance, in Figure \ref{app_fig:example_program2}, CodeT5 model misunderstands the problem and focuses on finding the greatest common divisor between $a$ and $b$ only.
Instead, the CodeRL model avoids this mistake and tackles the problem of finding the greatest common divisor between the \emph{factorials} of $a$ and $b$.

We also found that CodeRL can improve the complexity of the generated programs, an important quality in complex programming problems. 
For instance, in the interview-level program in Figure \ref{app_fig:example_program2}, we note that without applying CS, the generated program is functionally correct but fails during execution due to a timeout error.
This program simply computes separate factorials of both $a$ and $b$, which will slow down the execution in scenarios with extremely large $a$ or $b$. 
Applying the CS procedure can condition models on parts of the prior program and (re)generates new tokens to produce a more efficient program.
In the example in Figure \ref{app_fig:example_program2}, the factorials are computed on $\texttt{min(a,b)}$ to improve the efficiency of the programs. 
Hence, the resulting final program is able to pass all hidden unit tests (including tests with extremely large values) without timeout errors. 

For additional example generated programs, please refer to Appendix \ref{app_sec:example_programs}. 


\section{Limitations and Broader Impacts}\label{sec:limitation}
Program synthesis can lead to substantial positive social impacts, e.g., transforming future software developing tools, increasing the productivity of developers, and improving the accessibility and quality of programming courses. We propose CodeRL as a general framework to improve the performance of code generated from language models. We refer readers to the limitations and broader impacts discussed at length by \cite{chen2021evaluating} as these would apply to the different actor models one can use with CodeRL. The risks and limitations are critical to be considered before deploying such models at scale. 

One major limitation highlighted in many contemporary systems is that language models trained on code can generate code that is biased, and can even generate toxic natural language as code comments. Similar to previous work in language generation \citep{ouyang2022training}, besides improving functional correctness, RL could be used to align the models as per human preferences. Guided or weighted decoding schemes \citep{krause2020gedi} or safety-specific control tokens \citep{xu2020recipes} could also be explored to guide the generation of code towards desirable attributes like being secure, reliable, fast, efficient, fair, and representative.

Another data bias we need to consider is related to the system security of pretraining data. 
For example, pretraining data from public Github code repositories may contain vulnerabilities and the resulting synthesis models may generate programs with weak security measures \citep{hammond2021empirical}. Thus, similar to other code generation systems based on large LMs, CodeRL's output does warrant verification by qualified human developers.

Another limitation of our approach is the computational cost of training the critic model to estimate returns in addition to the original LM (actor network). However, in practice, we found that training a good critic model does not require large-scale models to attain a decent performance. For instance, a finetuned critic model initialized from a pretrained GPT-2 (small) can achieve over $75\%$ error prediction accuracy on synthetic samples. Thus fine-tuning costs for the critic model are minor compared to the pretraining of the original LM. With CodeRL, we obtain performance superior to much larger pre-trained language models.
Combining CodeRL with smaller models (e.g. a CodeT5-770M) will have significantly lesser inference costs for generating programs.  

Finally, previous works on code generation have highlighted how systems trained with the next token prediction objective exhibit alignment failure - model outputs not being aligned with the user's intent despite the model being capable of doing so. This holds true for CodeRL too, as we do witness generated code that sometimes does not satisfy user requirements in terms of the unit tests. However, unlike previous works, CodeRL's output can be tailored heavily by a user through the problem description as well as the unit tests that the solution is expected to pass. By leveraging unit tests during training, and during inference, CodeRL, when applied with a base code generation system, improves its alignment with the user intent. CodeRL's ability to improve alignment can be crucial in addressing misalignment issues which are predicted by \citet{chen2021evaluating} to get worse as we scale up data, parameters, and training time. 

\section{Conclusion}

We present CodeRL, a novel framework for program synthesis, using deep reinforcement learning to improve pretrained LMs, by exploiting unit test signals in both training and inference stages. Specifically, we introduce an actor-critic training approach to optimize pretrained LMs with dense feedback signals on synthetic code samples. During inference, we propose a new generation procedure with critical sampling, which enables the model to automatically regenerate programs based on feedback from unit tests and critic scores.
We integrate CodeRL with the improved CodeT5-large model (770M) and achieve new SOTA results on both the APPS and MBPP benchmarks, surpassing the prior SOTA by massive pretrained LMs of much larger model sizes. Our comprehensive analysis shows that CodeRL achieved consistent improvement upon the conventional pretrained LMs for code generation tasks. CodeRL is a general framework that integrates pretrained LMs and RL holistically for program synthesis, and can be extended and improved in various ways. For example, it can be easily integrated with other better pretrained LMs and can be improved with more fine-grained feedback from the environment, such as feedback received from a static code analyzer.

\section*{Acknowledgements}
We would like to thank our Salesforce Research team for fruitful discussions and feedback. 


\bibliography{neurips}

\begin{thebibliography}{76}
\providecommand{\natexlab}[1]{#1}
\providecommand{\url}[1]{\texttt{#1}}
\expandafter\ifx\csname urlstyle\endcsname\relax
  \providecommand{\doi}[1]{doi: #1}\else
  \providecommand{\doi}{doi: \begingroup \urlstyle{rm}\Url}\fi

\bibitem[Austin et~al.(2021)Austin, Odena, Nye, Bosma, Michalewski, Dohan,
  Jiang, Cai, Terry, Le, et~al.]{austin2021program}
J.~Austin, A.~Odena, M.~Nye, M.~Bosma, H.~Michalewski, D.~Dohan, E.~Jiang,
  C.~Cai, M.~Terry, Q.~Le, et~al.
\newblock Program synthesis with large language models.
\newblock \emph{arXiv preprint arXiv:2108.07732}, 2021.

\bibitem[Aye et~al.(2021)Aye, Kim, and Li]{aye2021learning}
G.~A. Aye, S.~Kim, and H.~Li.
\newblock Learning autocompletion from real-world datasets.
\newblock In \emph{2021 IEEE/ACM 43rd International Conference on Software
  Engineering: Software Engineering in Practice (ICSE-SEIP)}, pages 131--139.
  IEEE, 2021.

\bibitem[Bahdanau et~al.(2016)Bahdanau, Brakel, Xu, Goyal, Lowe, Pineau,
  Courville, and Bengio]{bahdanau2016actor}
D.~Bahdanau, P.~Brakel, K.~Xu, A.~Goyal, R.~Lowe, J.~Pineau, A.~Courville, and
  Y.~Bengio.
\newblock An actor-critic algorithm for sequence prediction.
\newblock \emph{arXiv preprint arXiv:1607.07086}, 2016.

\bibitem[Balog et~al.(2016)Balog, Gaunt, Brockschmidt, Nowozin, and
  Tarlow]{balog2016deepcoder}
M.~Balog, A.~L. Gaunt, M.~Brockschmidt, S.~Nowozin, and D.~Tarlow.
\newblock Deepcoder: Learning to write programs.
\newblock \emph{arXiv preprint arXiv:1611.01989}, 2016.

\bibitem[Bengio et~al.(2015)Bengio, Vinyals, Jaitly, and
  Shazeer]{bengio2015scheduled}
S.~Bengio, O.~Vinyals, N.~Jaitly, and N.~Shazeer.
\newblock Scheduled sampling for sequence prediction with recurrent neural
  networks.
\newblock \emph{Advances in neural information processing systems}, 28, 2015.

\bibitem[Black et~al.(2021)Black, Leo, Wang, Leahy, and Biderman]{black10gpt}
S.~Black, G.~Leo, P.~Wang, C.~Leahy, and S.~Biderman.
\newblock Gpt-neo: Large scale autoregressive language modeling with
  mesh-tensorflow.
\newblock \emph{URL https://doi. org/10.5281/zenodo}, 5297715, 2021.

\bibitem[Brown et~al.(2020)Brown, Mann, Ryder, Subbiah, Kaplan, Dhariwal,
  Neelakantan, Shyam, Sastry, Askell, et~al.]{brown2020language}
T.~Brown, B.~Mann, N.~Ryder, M.~Subbiah, J.~D. Kaplan, P.~Dhariwal,
  A.~Neelakantan, P.~Shyam, G.~Sastry, A.~Askell, et~al.
\newblock Language models are few-shot learners.
\newblock \emph{Advances in neural information processing systems},
  33:\penalty0 1877--1901, 2020.

\bibitem[Bruch et~al.(2009)Bruch, Monperrus, and Mezini]{bruch2009learning}
M.~Bruch, M.~Monperrus, and M.~Mezini.
\newblock Learning from examples to improve code completion systems.
\newblock In \emph{Proceedings of the 7th joint meeting of the European
  software engineering conference and the ACM SIGSOFT symposium on the
  foundations of software engineering}, pages 213--222, 2009.

\bibitem[Bunel et~al.(2018)Bunel, Hausknecht, Devlin, Singh, and
  Kohli]{bunel2018leveraging}
R.~Bunel, M.~Hausknecht, J.~Devlin, R.~Singh, and P.~Kohli.
\newblock Leveraging grammar and reinforcement learning for neural program
  synthesis.
\newblock In \emph{International Conference on Learning Representations}, 2018.
\newblock URL \url{https://openreview.net/forum?id=H1Xw62kRZ}.

\bibitem[Chen et~al.(2021{\natexlab{a}})Chen, Tworek, Jun, Yuan, Pinto, Kaplan,
  Edwards, Burda, Joseph, Brockman, et~al.]{chen2021evaluating}
M.~Chen, J.~Tworek, H.~Jun, Q.~Yuan, H.~P. d.~O. Pinto, J.~Kaplan, H.~Edwards,
  Y.~Burda, N.~Joseph, G.~Brockman, et~al.
\newblock Evaluating large language models trained on code.
\newblock \emph{arXiv preprint arXiv:2107.03374}, 2021{\natexlab{a}}.

\bibitem[Chen et~al.(2022)Chen, Lacomis, Schwartz, Neubig, Vasilescu, and
  {Le~Goues}]{ChenVarCLR2022}
Q.~Chen, J.~Lacomis, E.~J. Schwartz, G.~Neubig, B.~Vasilescu, and
  C.~{Le~Goues}.
\newblock {VarCLR}: {Variable} semantic representation pre-training via
  contrastive learning.
\newblock In \emph{International Conference on Software Engineering}, ICSE '22,
  2022.

\bibitem[Chen et~al.(2021{\natexlab{b}})Chen, Song, and Tian]{chen2021latent}
X.~Chen, D.~Song, and Y.~Tian.
\newblock Latent execution for neural program synthesis beyond domain-specific
  languages.
\newblock \emph{Advances in Neural Information Processing Systems}, 34,
  2021{\natexlab{b}}.

\bibitem[Clement et~al.(2020)Clement, Drain, Timcheck, Svyatkovskiy, and
  Sundaresan]{clement-etal-2020-pymt5}
C.~Clement, D.~Drain, J.~Timcheck, A.~Svyatkovskiy, and N.~Sundaresan.
\newblock {P}y{MT}5: multi-mode translation of natural language and python code
  with transformers.
\newblock In \emph{Proceedings of the 2020 Conference on Empirical Methods in
  Natural Language Processing (EMNLP)}, pages 9052--9065, Online, Nov. 2020.
  Association for Computational Linguistics.
\newblock \doi{10.18653/v1/2020.emnlp-main.728}.
\newblock URL \url{https://aclanthology.org/2020.emnlp-main.728}.

\bibitem[Cobbe et~al.(2021)Cobbe, Kosaraju, Bavarian, Hilton, Nakano, Hesse,
  and Schulman]{cobbe2021training}
K.~Cobbe, V.~Kosaraju, M.~Bavarian, J.~Hilton, R.~Nakano, C.~Hesse, and
  J.~Schulman.
\newblock Training verifiers to solve math word problems.
\newblock \emph{arXiv preprint arXiv:2110.14168}, 2021.

\bibitem[Devlin et~al.(2017)Devlin, Uesato, Bhupatiraju, Singh, Mohamed, and
  Kohli]{devlin2017robustfill}
J.~Devlin, J.~Uesato, S.~Bhupatiraju, R.~Singh, A.-r. Mohamed, and P.~Kohli.
\newblock Robustfill: Neural program learning under noisy i/o.
\newblock In \emph{International conference on machine learning}, pages
  990--998. PMLR, 2017.

\bibitem[Ellis et~al.(2018)Ellis, Ritchie, Solar-Lezama, and
  Tenenbaum]{ellis2018learning}
K.~Ellis, D.~Ritchie, A.~Solar-Lezama, and J.~Tenenbaum.
\newblock Learning to infer graphics programs from hand-drawn images.
\newblock \emph{Advances in neural information processing systems}, 31, 2018.

\bibitem[Ellis et~al.(2019)Ellis, Nye, Pu, Sosa, Tenenbaum, and
  Solar-Lezama]{ellis2019write}
K.~Ellis, M.~Nye, Y.~Pu, F.~Sosa, J.~Tenenbaum, and A.~Solar-Lezama.
\newblock Write, execute, assess: Program synthesis with a repl.
\newblock \emph{Advances in Neural Information Processing Systems}, 32, 2019.

\bibitem[Feng et~al.(2020)Feng, Guo, Tang, Duan, Feng, Gong, Shou, Qin, Liu,
  Jiang, and Zhou]{feng-etal-2020-codebert}
Z.~Feng, D.~Guo, D.~Tang, N.~Duan, X.~Feng, M.~Gong, L.~Shou, B.~Qin, T.~Liu,
  D.~Jiang, and M.~Zhou.
\newblock {C}ode{BERT}: A pre-trained model for programming and natural
  languages.
\newblock In \emph{Findings of the Association for Computational Linguistics:
  EMNLP 2020}, pages 1536--1547, Online, Nov. 2020. Association for
  Computational Linguistics.
\newblock \doi{10.18653/v1/2020.findings-emnlp.139}.
\newblock URL \url{https://aclanthology.org/2020.findings-emnlp.139}.

\bibitem[Ganin et~al.(2018)Ganin, Kulkarni, Babuschkin, Eslami, and
  Vinyals]{ganin2018synthesizing}
Y.~Ganin, T.~Kulkarni, I.~Babuschkin, S.~A. Eslami, and O.~Vinyals.
\newblock Synthesizing programs for images using reinforced adversarial
  learning.
\newblock In \emph{International Conference on Machine Learning}, pages
  1666--1675. PMLR, 2018.

\bibitem[Gulwani et~al.(2012)Gulwani, Harris, and
  Singh]{gulwani2012spreadsheet}
S.~Gulwani, W.~R. Harris, and R.~Singh.
\newblock Spreadsheet data manipulation using examples.
\newblock \emph{Communications of the ACM}, 55\penalty0 (8):\penalty0 97--105,
  2012.

\bibitem[Guo et~al.(2021)Guo, Svyatkovskiy, Yin, Duan, Brockschmidt, and
  Allamanis]{guo2021learning}
D.~Guo, A.~Svyatkovskiy, J.~Yin, N.~Duan, M.~Brockschmidt, and M.~Allamanis.
\newblock Learning to complete code with sketches.
\newblock In \emph{International Conference on Learning Representations}, 2021.

\bibitem[Guu et~al.(2017)Guu, Pasupat, Liu, and Liang]{guu-etal-2017-language}
K.~Guu, P.~Pasupat, E.~Liu, and P.~Liang.
\newblock From language to programs: Bridging reinforcement learning and
  maximum marginal likelihood.
\newblock In \emph{Proceedings of the 55th Annual Meeting of the Association
  for Computational Linguistics (Volume 1: Long Papers)}, pages 1051--1062,
  Vancouver, Canada, July 2017. Association for Computational Linguistics.
\newblock \doi{10.18653/v1/P17-1097}.
\newblock URL \url{https://aclanthology.org/P17-1097}.

\bibitem[Hammond~Pearce et~al.(2021)Hammond~Pearce, Tan, Dolan-Gavitt, and
  Karri]{hammond2021empirical}
B.~A. Hammond~Pearce, B.~Tan, B.~Dolan-Gavitt, and R.~Karri.
\newblock An empirical cybersecurity evaluation of github copilot’s code
  contributions.
\newblock \emph{arXiv preprint arXiv:2108.09293}, 2021.

\bibitem[Hendrycks et~al.(2021)Hendrycks, Basart, Kadavath, Mazeika, Arora,
  Guo, Burns, Puranik, He, Song, and Steinhardt]{hendrycksapps2021}
D.~Hendrycks, S.~Basart, S.~Kadavath, M.~Mazeika, A.~Arora, E.~Guo, C.~Burns,
  S.~Puranik, H.~He, D.~Song, and J.~Steinhardt.
\newblock Measuring coding challenge competence with apps.
\newblock \emph{NeurIPS}, 2021.

\bibitem[Husain et~al.(2019)Husain, Wu, Gazit, Allamanis, and
  Brockschmidt]{csn}
H.~Husain, H.~Wu, T.~Gazit, M.~Allamanis, and M.~Brockschmidt.
\newblock Codesearchnet challenge: Evaluating the state of semantic code
  search.
\newblock \emph{CoRR}, abs/1909.09436, 2019.

\bibitem[Johnson et~al.(2017)Johnson, Hariharan, Van Der~Maaten, Hoffman,
  Fei-Fei, Lawrence~Zitnick, and Girshick]{johnson2017inferring}
J.~Johnson, B.~Hariharan, L.~Van Der~Maaten, J.~Hoffman, L.~Fei-Fei,
  C.~Lawrence~Zitnick, and R.~Girshick.
\newblock Inferring and executing programs for visual reasoning.
\newblock In \emph{Proceedings of the IEEE International Conference on Computer
  Vision}, pages 2989--2998, 2017.

\bibitem[Joulin and Mikolov(2015)]{joulin2015inferring}
A.~Joulin and T.~Mikolov.
\newblock Inferring algorithmic patterns with stack-augmented recurrent nets.
\newblock \emph{Advances in neural information processing systems}, 28, 2015.

\bibitem[Konda and Tsitsiklis(1999)]{konda1999actor}
V.~Konda and J.~Tsitsiklis.
\newblock Actor-critic algorithms.
\newblock \emph{Advances in neural information processing systems}, 12, 1999.

\bibitem[Krause et~al.(2021)Krause, Gotmare, McCann, Keskar, Joty, Socher, and
  Rajani]{krause2020gedi}
B.~Krause, A.~D. Gotmare, B.~McCann, N.~S. Keskar, S.~Joty, R.~Socher, and
  N.~F. Rajani.
\newblock {G}e{D}i: Generative discriminator guided sequence generation.
\newblock In \emph{Findings of the Association for Computational Linguistics:
  EMNLP 2021}, pages 4929--4952, Punta Cana, Dominican Republic, Nov. 2021.
  Association for Computational Linguistics.
\newblock \doi{10.18653/v1/2021.findings-emnlp.424}.
\newblock URL \url{https://aclanthology.org/2021.findings-emnlp.424}.

\bibitem[Kulkarni et~al.(2015)Kulkarni, Whitney, Kohli, and
  Tenenbaum]{kulkarni2015deep}
T.~D. Kulkarni, W.~F. Whitney, P.~Kohli, and J.~Tenenbaum.
\newblock Deep convolutional inverse graphics network.
\newblock \emph{Advances in neural information processing systems}, 28, 2015.

\bibitem[Kurach et~al.(2015)Kurach, Andrychowicz, and
  Sutskever]{kurach2015neural}
K.~Kurach, M.~Andrychowicz, and I.~Sutskever.
\newblock Neural random-access machines.
\newblock \emph{arXiv preprint arXiv:1511.06392}, 2015.

\bibitem[Li et~al.(2022)Li, Choi, Chung, Kushman, Schrittwieser, Leblond,
  Eccles, Keeling, Gimeno, Lago, et~al.]{li2022competition}
Y.~Li, D.~Choi, J.~Chung, N.~Kushman, J.~Schrittwieser, R.~Leblond, T.~Eccles,
  J.~Keeling, F.~Gimeno, A.~D. Lago, et~al.
\newblock Competition-level code generation with alphacode.
\newblock \emph{arXiv preprint arXiv:2203.07814}, 2022.

\bibitem[Liang et~al.(2018)Liang, Norouzi, Berant, Le, and
  Lao]{liang2018memory}
C.~Liang, M.~Norouzi, J.~Berant, Q.~V. Le, and N.~Lao.
\newblock Memory augmented policy optimization for program synthesis and
  semantic parsing.
\newblock \emph{Advances in Neural Information Processing Systems}, 31, 2018.

\bibitem[Liang et~al.(2010)Liang, Jordan, and Klein]{liang2010learning}
P.~Liang, M.~I. Jordan, and D.~Klein.
\newblock Learning programs: A hierarchical bayesian approach.
\newblock In \emph{Proceedings of the 27th International Conference on Machine
  Learning (ICML-10)}, pages 639--646, 2010.

\bibitem[Lillicrap et~al.(2015)Lillicrap, Hunt, Pritzel, Heess, Erez, Tassa,
  Silver, and Wierstra]{lillicrap2015continuous}
T.~P. Lillicrap, J.~J. Hunt, A.~Pritzel, N.~Heess, T.~Erez, Y.~Tassa,
  D.~Silver, and D.~Wierstra.
\newblock Continuous control with deep reinforcement learning.
\newblock \emph{arXiv preprint arXiv:1509.02971}, 2015.

\bibitem[Lin(2004)]{lin2004rouge}
C.-Y. Lin.
\newblock Rouge: A package for automatic evaluation of summaries.
\newblock \emph{Text Summarization Branches Out}, 2004.

\bibitem[Liu et~al.(2019)Liu, Wu, Wu, Ritchie, Freeman, and
  Tenenbaum]{liu2018learning}
Y.~Liu, J.~Wu, Z.~Wu, D.~Ritchie, W.~T. Freeman, and J.~B. Tenenbaum.
\newblock Learning to describe scenes with programs.
\newblock In \emph{International Conference on Learning Representations}, 2019.
\newblock URL \url{https://openreview.net/forum?id=SyNPk2R9K7}.

\bibitem[Loshchilov and Hutter(2019)]{DBLP:conf/iclr/LoshchilovH19}
I.~Loshchilov and F.~Hutter.
\newblock Decoupled weight decay regularization.
\newblock In \emph{{ICLR} (Poster)}. OpenReview.net, 2019.

\bibitem[Lu et~al.(2021)Lu, Guo, Ren, Huang, Svyatkovskiy, Blanco, Clement,
  Drain, Jiang, Tang, Li, Zhou, Shou, Zhou, Tufano, Gong, Zhou, Duan,
  Sundaresan, Deng, Fu, and Liu]{codexglue}
S.~Lu, D.~Guo, S.~Ren, J.~Huang, A.~Svyatkovskiy, A.~Blanco, C.~B. Clement,
  D.~Drain, D.~Jiang, D.~Tang, G.~Li, L.~Zhou, L.~Shou, L.~Zhou, M.~Tufano,
  M.~Gong, M.~Zhou, N.~Duan, N.~Sundaresan, S.~K. Deng, S.~Fu, and S.~Liu.
\newblock Codexglue: {A} machine learning benchmark dataset for code
  understanding and generation.
\newblock In \emph{NeurIPS Datasets and Benchmarks}, 2021.

\bibitem[Manna and Waldinger(1971)]{manna1971toward}
Z.~Manna and R.~J. Waldinger.
\newblock Toward automatic program synthesis.
\newblock \emph{Communications of the ACM}, 14\penalty0 (3):\penalty0 151--165,
  1971.

\bibitem[Nijkamp et~al.(2022)Nijkamp, Pang, Hayashi, Tu, Wang, Zhou, Savarese,
  and Xiong]{nijkamp2022conversational}
E.~Nijkamp, B.~Pang, H.~Hayashi, L.~Tu, H.~Wang, Y.~Zhou, S.~Savarese, and
  C.~Xiong.
\newblock A conversational paradigm for program synthesis.
\newblock \emph{arXiv preprint arXiv:2203.13474}, 2022.

\bibitem[Ouyang et~al.(2022)Ouyang, Wu, Jiang, Almeida, Wainwright, Mishkin,
  Zhang, Agarwal, Slama, Ray, et~al.]{ouyang2022training}
L.~Ouyang, J.~Wu, X.~Jiang, D.~Almeida, C.~L. Wainwright, P.~Mishkin, C.~Zhang,
  S.~Agarwal, K.~Slama, A.~Ray, et~al.
\newblock Training language models to follow instructions with human feedback.
\newblock \emph{arXiv preprint arXiv:2203.02155}, 2022.

\bibitem[Papineni et~al.(2002)Papineni, Roukos, Ward, and
  Zhu]{papineni2002bleu}
K.~Papineni, S.~Roukos, T.~Ward, and W.-J. Zhu.
\newblock Bleu: a method for automatic evaluation of machine translation.
\newblock In \emph{Proceedings of the 40th annual meeting on association for
  computational linguistics}, pages 311--318. Association for Computational
  Linguistics, 2002.

\bibitem[Parisotto et~al.(2016)Parisotto, Mohamed, Singh, Li, Zhou, and
  Kohli]{parisotto2016neuro}
E.~Parisotto, A.-r. Mohamed, R.~Singh, L.~Li, D.~Zhou, and P.~Kohli.
\newblock Neuro-symbolic program synthesis.
\newblock \emph{arXiv preprint arXiv:1611.01855}, 2016.

\bibitem[Poesia et~al.(2022)Poesia, Polozov, Le, Tiwari, Soares, Meek, and
  Gulwani]{poesia2022synchromesh}
G.~Poesia, A.~Polozov, V.~Le, A.~Tiwari, G.~Soares, C.~Meek, and S.~Gulwani.
\newblock Synchromesh: Reliable code generation from pre-trained language
  models.
\newblock In \emph{International Conference on Learning Representations}, 2022.
\newblock URL \url{https://openreview.net/forum?id=KmtVD97J43e}.

\bibitem[Radford et~al.(2019)Radford, Wu, Child, Luan, Amodei, Sutskever,
  et~al.]{radford2019language}
A.~Radford, J.~Wu, R.~Child, D.~Luan, D.~Amodei, I.~Sutskever, et~al.
\newblock Language models are unsupervised multitask learners.
\newblock \emph{OpenAI blog}, 1\penalty0 (8):\penalty0 9, 2019.

\bibitem[Raffel et~al.(2020)Raffel, Shazeer, Roberts, Lee, Narang, Matena,
  Zhou, Li, and Liu]{t5}
C.~Raffel, N.~Shazeer, A.~Roberts, K.~Lee, S.~Narang, M.~Matena, Y.~Zhou,
  W.~Li, and P.~J. Liu.
\newblock Exploring the limits of transfer learning with a unified text-to-text
  transformer.
\newblock \emph{J. Mach. Learn. Res.}, 21:\penalty0 140:1--140:67, 2020.

\bibitem[Ranzato et~al.(2016)Ranzato, Chopra, Auli, and
  Zaremba]{DBLP:journals/corr/RanzatoCAZ15}
M.~Ranzato, S.~Chopra, M.~Auli, and W.~Zaremba.
\newblock Sequence level training with recurrent neural networks.
\newblock In Y.~Bengio and Y.~LeCun, editors, \emph{4th International
  Conference on Learning Representations, {ICLR} 2016, San Juan, Puerto Rico,
  May 2-4, 2016, Conference Track Proceedings}, 2016.
\newblock URL \url{http://arxiv.org/abs/1511.06732}.

\bibitem[Raychev et~al.(2014)Raychev, Vechev, and Yahav]{raychev2014code}
V.~Raychev, M.~Vechev, and E.~Yahav.
\newblock Code completion with statistical language models.
\newblock In \emph{Proceedings of the 35th ACM SIGPLAN Conference on
  Programming Language Design and Implementation}, pages 419--428, 2014.

\bibitem[Ren et~al.(2020)Ren, Guo, Lu, Zhou, Liu, Tang, Sundaresan, Zhou,
  Blanco, and Ma]{ren2020codebleu}
S.~Ren, D.~Guo, S.~Lu, L.~Zhou, S.~Liu, D.~Tang, N.~Sundaresan, M.~Zhou,
  A.~Blanco, and S.~Ma.
\newblock Codebleu: a method for automatic evaluation of code synthesis.
\newblock \emph{arXiv preprint arXiv:2009.10297}, 2020.

\bibitem[Ren et~al.(2017)Ren, Wang, Zhang, Lv, and Li]{ren2017deep}
Z.~Ren, X.~Wang, N.~Zhang, X.~Lv, and L.-J. Li.
\newblock Deep reinforcement learning-based image captioning with embedding
  reward.
\newblock In \emph{Proceedings of the IEEE conference on computer vision and
  pattern recognition}, pages 290--298, 2017.

\bibitem[Rennie et~al.(2017)Rennie, Marcheret, Mroueh, Ross, and
  Goel]{rennie2017self}
S.~J. Rennie, E.~Marcheret, Y.~Mroueh, J.~Ross, and V.~Goel.
\newblock Self-critical sequence training for image captioning.
\newblock In \emph{Proceedings of the IEEE conference on computer vision and
  pattern recognition}, pages 7008--7024, 2017.

\bibitem[Robbes and Lanza(2008)]{robbes2008program}
R.~Robbes and M.~Lanza.
\newblock How program history can improve code completion.
\newblock In \emph{2008 23rd IEEE/ACM International Conference on Automated
  Software Engineering}, pages 317--326. IEEE, 2008.

\bibitem[Summers(1977)]{summers1977methodology}
P.~D. Summers.
\newblock A methodology for lisp program construction from examples.
\newblock \emph{Journal of the ACM (JACM)}, 24\penalty0 (1):\penalty0 161--175,
  1977.

\bibitem[Sun et~al.(2018)Sun, Noh, Somasundaram, and Lim]{pmlr-v80-sun18a}
S.-H. Sun, H.~Noh, S.~Somasundaram, and J.~Lim.
\newblock Neural program synthesis from diverse demonstration videos.
\newblock In J.~Dy and A.~Krause, editors, \emph{Proceedings of the 35th
  International Conference on Machine Learning}, volume~80 of \emph{Proceedings
  of Machine Learning Research}, pages 4790--4799. PMLR, 10--15 Jul 2018.
\newblock URL \url{https://proceedings.mlr.press/v80/sun18a.html}.

\bibitem[Sutton(1984)]{sutton1984temporal}
R.~S. Sutton.
\newblock \emph{Temporal credit assignment in reinforcement learning}.
\newblock PhD thesis, University of Massachusetts Amherst, 1984.

\bibitem[Sutton and Barto(2018)]{sutton2018reinforcement}
R.~S. Sutton and A.~G. Barto.
\newblock \emph{Reinforcement learning: An introduction}.
\newblock MIT press, 2018.

\bibitem[Sutton et~al.(1999)Sutton, McAllester, Singh, and
  Mansour]{sutton1999policy}
R.~S. Sutton, D.~McAllester, S.~Singh, and Y.~Mansour.
\newblock Policy gradient methods for reinforcement learning with function
  approximation.
\newblock \emph{Advances in neural information processing systems}, 12, 1999.

\bibitem[Svyatkovskiy et~al.(2020)Svyatkovskiy, Deng, Fu, and
  Sundaresan]{svyatkovskiy2020intellicode}
A.~Svyatkovskiy, S.~K. Deng, S.~Fu, and N.~Sundaresan.
\newblock Intellicode compose: Code generation using transformer.
\newblock In \emph{Proceedings of the 28th ACM Joint Meeting on European
  Software Engineering Conference and Symposium on the Foundations of Software
  Engineering}, pages 1433--1443, 2020.

\bibitem[Svyatkovskiy et~al.(2021)Svyatkovskiy, Lee, Hadjitofi, Riechert,
  Franco, and Allamanis]{svyatkovskiy2021fast}
A.~Svyatkovskiy, S.~Lee, A.~Hadjitofi, M.~Riechert, J.~V. Franco, and
  M.~Allamanis.
\newblock Fast and memory-efficient neural code completion.
\newblock In \emph{2021 IEEE/ACM 18th International Conference on Mining
  Software Repositories (MSR)}, pages 329--340. IEEE, 2021.

\bibitem[Tian et~al.(2019)Tian, Luo, Sun, Ellis, Freeman, Tenenbaum, and
  Wu]{tian2018learning}
Y.~Tian, A.~Luo, X.~Sun, K.~Ellis, W.~T. Freeman, J.~B. Tenenbaum, and J.~Wu.
\newblock Learning to infer and execute 3d shape programs.
\newblock In \emph{International Conference on Learning Representations}, 2019.
\newblock URL \url{https://openreview.net/forum?id=rylNH20qFQ}.

\bibitem[Trivedi et~al.(2021)Trivedi, Zhang, Sun, and Lim]{trivedi2021learning}
D.~Trivedi, J.~Zhang, S.-H. Sun, and J.~J. Lim.
\newblock Learning to synthesize programs as interpretable and generalizable
  policies.
\newblock \emph{Advances in Neural Information Processing Systems},
  34:\penalty0 25146--25163, 2021.

\bibitem[Vaswani et~al.(2017)Vaswani, Shazeer, Parmar, Uszkoreit, Jones, Gomez,
  Kaiser, and Polosukhin]{vaswani2017attention}
A.~Vaswani, N.~Shazeer, N.~Parmar, J.~Uszkoreit, L.~Jones, A.~N. Gomez,
  {\L}.~Kaiser, and I.~Polosukhin.
\newblock Attention is all you need.
\newblock \emph{Advances in neural information processing systems}, 30, 2017.

\bibitem[Vedantam et~al.(2015)Vedantam, Lawrence~Zitnick, and
  Parikh]{vedantam2015cider}
R.~Vedantam, C.~Lawrence~Zitnick, and D.~Parikh.
\newblock Cider: Consensus-based image description evaluation.
\newblock In \emph{Proceedings of the IEEE conference on computer vision and
  pattern recognition}, pages 4566--4575, 2015.

\bibitem[Waldinger and Lee(1969)]{waldinger1969prow}
R.~J. Waldinger and R.~C. Lee.
\newblock Prow: A step toward automatic program writing.
\newblock In \emph{Proceedings of the 1st international joint conference on
  Artificial intelligence}, pages 241--252, 1969.

\bibitem[Wang and Komatsuzaki(2021)]{gpt-j}
B.~Wang and A.~Komatsuzaki.
\newblock {GPT-J-6B: A 6 Billion Parameter Autoregressive Language Model}.
\newblock \url{https://github.com/kingoflolz/mesh-transformer-jax}, May 2021.

\bibitem[Wang et~al.(2018)Wang, Chen, Wu, Wang, and Wang]{wang2018video}
X.~Wang, W.~Chen, J.~Wu, Y.-F. Wang, and W.~Y. Wang.
\newblock Video captioning via hierarchical reinforcement learning.
\newblock In \emph{Proceedings of the IEEE Conference on Computer Vision and
  Pattern Recognition}, pages 4213--4222, 2018.

\bibitem[Wang et~al.(2021)Wang, Wang, Joty, and Hoi]{codet5}
Y.~Wang, W.~Wang, S.~R. Joty, and S.~C.~H. Hoi.
\newblock Codet5: Identifier-aware unified pre-trained encoder-decoder models
  for code understanding and generation.
\newblock In \emph{{EMNLP} {(1)}}, pages 8696--8708. Association for
  Computational Linguistics, 2021.

\bibitem[White et~al.(2015)White, Vendome, Linares-V{\'a}squez, and
  Poshyvanyk]{white2015toward}
M.~White, C.~Vendome, M.~Linares-V{\'a}squez, and D.~Poshyvanyk.
\newblock Toward deep learning software repositories.
\newblock In \emph{2015 IEEE/ACM 12th Working Conference on Mining Software
  Repositories}, pages 334--345. IEEE, 2015.

\bibitem[Williams(1992)]{williams1992simple}
R.~J. Williams.
\newblock Simple statistical gradient-following algorithms for connectionist
  reinforcement learning.
\newblock \emph{Machine learning}, 8\penalty0 (3):\penalty0 229--256, 1992.

\bibitem[Wu et~al.(2017)Wu, Tenenbaum, and Kohli]{Wu_2017_CVPR}
J.~Wu, J.~B. Tenenbaum, and P.~Kohli.
\newblock Neural scene de-rendering.
\newblock In \emph{Proceedings of the IEEE Conference on Computer Vision and
  Pattern Recognition (CVPR)}, July 2017.

\bibitem[Xu et~al.(2020)Xu, Ju, Li, Boureau, Weston, and Dinan]{xu2020recipes}
J.~Xu, D.~Ju, M.~Li, Y.-L. Boureau, J.~Weston, and E.~Dinan.
\newblock Recipes for safety in open-domain chatbots.
\newblock \emph{arXiv preprint arXiv:2010.07079}, 2020.

\bibitem[Xu et~al.(2018)Xu, Liu, and Song]{xu2018sqlnet}
X.~Xu, C.~Liu, and D.~Song.
\newblock {SQLN}et: Generating structured queries from natural language without
  reinforcement learning, 2018.
\newblock URL \url{https://openreview.net/forum?id=SkYibHlRb}.

\bibitem[Yang et~al.(2015)Yang, Reed, Yang, and Lee]{yang2015weakly}
J.~Yang, S.~E. Reed, M.-H. Yang, and H.~Lee.
\newblock Weakly-supervised disentangling with recurrent transformations for 3d
  view synthesis.
\newblock \emph{Advances in neural information processing systems}, 28, 2015.

\bibitem[Yin and Neubig(2017)]{yin-neubig-2017-syntactic}
P.~Yin and G.~Neubig.
\newblock A syntactic neural model for general-purpose code generation.
\newblock In \emph{Proceedings of the 55th Annual Meeting of the Association
  for Computational Linguistics (Volume 1: Long Papers)}, pages 440--450,
  Vancouver, Canada, July 2017. Association for Computational Linguistics.
\newblock \doi{10.18653/v1/P17-1041}.
\newblock URL \url{https://aclanthology.org/P17-1041}.

\bibitem[Zhong et~al.(2018)Zhong, Xiong, and Socher]{zhong2018seqsql}
V.~Zhong, C.~Xiong, and R.~Socher.
\newblock Seq2{SQL}: Generating structured queries from natural language using
  reinforcement learning, 2018.
\newblock URL \url{https://openreview.net/forum?id=Syx6bz-Ab}.

\end{thebibliography}
\bibliographystyle{abbrvnat}


\appendix

\section{Critic Sampling Procedure}
\label{app_sec:critic_sampling}
\begin{algorithm}[htbp]
\SetNoFillComment
\DontPrintSemicolon
\KwIn{
Problem $D$, Language model $\theta$, Critic model $\phi$, Language model for program repair $\omega$}
\KwOut{A set of $N$ generated solution programs}
\Begin{

\tcc{extract example unit tests from problem}
\bf{example unit tests}\ $\mathcal{I} \longleftarrow \mathrm{ExtractExampleInputOutput} (D)$

\bf{a set of programs for repair} $\mathcal{R} \leftarrow \emptyset$

\While{$\mathrm{True}$}{
\bf{a set of output programs} $\mathcal{S} \longleftarrow \emptyset$

\If{$|\mathcal{R}|=0$}{
\tcc{generate program solutions by LM}

\For{$N$ {\bf times}}{
$\hat{W}_i \longleftarrow \mathrm{Sampling}\ p_\theta (W_i| D)
\longleftarrow \mathrm{Sampling}\ p_\theta (w^i_t| w^i_{1:t-1}, D)$

$\mathcal{S} \longleftarrow \mathcal{S} \cup \{\hat{W}_i\}$

}
}
\Else{
\tcc{repair current programs by LM}

{upsampling param} $K = N/ |\mathcal{R}|$

\For{{\bf each filtered program} $\hat{W}^{\mathrm{fail}}_i \in \mathcal{R}$}{
Obtain corresponding test outcome $u_i$ and error message $c_i$

\For{$K$ {\bf times}}{
$\hat{W}_j \longleftarrow \mathrm{Sampling}\ p_\omega (W_j| D, \hat{W}^{\mathrm{fail}}_i, u_i, c_i)
\longleftarrow \mathrm{Sampling}\ p_\omega (w^j_t| w^j_{1:t-1}, D, \hat{W}^{\mathrm{fail}}_i, u_i, c_i)$

$\mathcal{S} \longleftarrow \mathcal{S} \cup \{\hat{W}_j\}$

}
}
}

\tcc{filter programs based on their test results}

\bf{sets of filtered programs} $\mathcal{P} \leftarrow \emptyset$, $\mathcal{F} \leftarrow \emptyset$

\For{{\bf each generated program} $\hat{W}_i \in \mathcal{S}$}{
${\bf Test outcomes}\ {u_i} \longleftarrow \mathrm{RunTests}(\hat{W}_i, \mathcal{I})$

\lIf{$u_i = \mathrm{``PassedTest"}$}{
$\mathcal{P} \longleftarrow \mathcal{P} \cup \{\hat{W}_i\} $
}
\lElse{
$\mathcal{F} \longleftarrow \mathcal{F} \cup \{\hat{W}_i\} $
}
}

\lIf{$|\mathcal{P}|>0$}{
break
}
\Else{
\tcc{Sort failed programs and select top candidates}
Sort $\mathcal{F}$ by critic scores $p_\phi(u_i=\mathrm{PassedTest}|\hat{W}_i, D)\ \forall\ \hat{W}_i \in \mathcal{F}$

\bf{a set of programs} $\mathcal{R} \longleftarrow$ top-$M$ from $\mathcal{F}$

}
}

\tcc{Use passed programs to select sub-sequences}

{upsampling param} $N^{'} = N/ |\mathcal{P}|$

a set of output solutions $\mathcal{S}^{'} \longleftarrow \emptyset$

\For{{\bf each filtered program} $\hat{W}_i \in \mathcal{P}$}{
subsequence $\hat{W}^{sub}_i  \longleftarrow \mathrm{Sampling}\ p_\phi(u_i=\mathrm{PassedTest}|\hat{W}_i, D)$
    
    $\qquad \longleftarrow \mathrm{Sampling}\ p_\phi(u_i=\mathrm{PassedTest}|\hat{w}_{1:t-1}, D)$
    
    \tcc{subsequence as seeds to regenerate programs}
    
    length of subsequence $m \longleftarrow |\hat{W}^{sub}_i|$
    
    \For{$N^{'}$ {\bf times}}{
        $\hat{W}_j \longleftarrow \mathrm{Sampling}\ p_\theta (W_j| \hat{W}^{sub}_i, D)
\longleftarrow \mathrm{Sampling}\ p_\theta (w^j_t| w^j_{m:t-1}, \hat{W}^{sub}_i, D)$

        $\mathcal{S}^{'} \longleftarrow \mathcal{S}^{'} \cup \{\hat{W}_j\}$
    }
}
\tcc{return the regenerated programs for evaluation}
\Return{$\mathcal{S}^{'}$}
}
\caption{Critic sampling procedure to generate programs}
\label{algo:critic_sampling}
\end{algorithm}

Refer to Algorithm \ref{algo:critic_sampling} for a step-by-step explanation of our critic sampling procedure.  

\section{Additional Experimental Results}
\label{app_sec:exps}

\subsection{CodeXGLUE Benchmark Results}
\label{app_sec:codexglue}

To validate the effectiveness of our simplified pretraining strategies of CodeT5-large, we extensively evaluate it on a variety of generation tasks in CodeXGLUE~\citep{codexglue}, including code-to-text generation (i.e. summarization, see Table~\ref{app_tab:codexglue_sum}), text-to-code generation (see Table~\ref{app_tab:codexglue_gen}),  and code-to-code generation (i.e., code translation and code refinement, see Table~\ref{app_tab:codexglue_code2code}). 
Different from APPS \citep{hendrycksapps2021} and MBPP \citep{austin2021program}, we follow the default similarity-based evaluation metrics in the CodeXGLUE benchmark, including BLEU \citep{papineni2002bleu} and CodeBLEU \citep{ren2020codebleu}, and exact match (EM) scores. 
Table \ref{app_tab:codexglue_sum}, \ref{app_tab:codexglue_gen}, and \ref{app_tab:codexglue_code2code} show that our simplified pretrained CodeT5-large sets new SOTA results on a large majority of the tasks, and hence, can be served as a better foundation model for other code-related generation tasks.
Note that in these experiments, 
we employ the conventional finetuning objective with $\mathcal{L}_{ce}$ and there might be potential to improve the performance further with our CodeRL framework. 

\subsection{MBPP Benchmark Results}
\label{app_sec:mbpp_results}

Following~\citet{austin2021program}, we adopt temperature sampling to generate multiple candidate solutions. We empirically find that CodeT5 benefits from
a higher temperature of $1.2$ (less greedy decoding or more diverse) than their GPT's temperature of $0.5$ on this benchmark.
 
\begin{table}[t]
\centering
\caption{Code-to-Text generation results (smoothed BLEU-4) on CodeXGLUE}
\label{app_tab:codexglue_sum}
\begin{tabular}{l|c c c c c c|c}
\hline
Model & Ruby & JavaScript & Go & Python & Java & PHP & Overall \\
\hline
RoBERTa  & 11.17 & 11.90 & 17.72 & 18.14 & 16.47 & 24.02 & 16.57 \\
CodeBERT & 12.16 & 14.90 & 18.07 & 19.06 & 17.65 & 25.16 & 17.83 \\ 
DOBF & -& -& -  &18.24 & 19.05 & - & - \\
PLBART &  14.11 &  15.56 &  18.91 & 19.30 & 18.45 & 23.58 & 18.32\\
CoTexT & 14.02	 &14.96	 &18.86	 &19.73	 &19.06	 &24.58	 & 18.55 \\
CodeT5-small & 14.87&	15.32&	19.25&	20.04&	19.92&	25.46&	19.14\\
CodeT5-base & 15.24&	16.16&	19.56&	20.01&	20.31&	26.03&	19.55\\
\hline
CodeT5-large &\textbf{15.58}	&\textbf{16.17}	&\textbf{19.69}	&\textbf{20.57}	&\textbf{20.74}&	\textbf{26.49}	&\textbf{19.87} \\
\hline
\end{tabular}
\end{table}
\begin{table}[t]
\centering
\caption{Text-to-Code generation results on CodeXGLUE}
\label{app_tab:codexglue_gen}
\begin{tabular}{l|ccc}
\hline
Model         & EM    & BLEU-4  & CodeBLEU \\
\hline
GPT-2           & 17.35 & 25.37 & 29.69    \\
CodeGPT-2       & 18.25 & 28.69 & 32.71    \\
CodeGPT-adapted & 20.10 & 32.79 & 35.98    \\
PLBART          & 18.75 & 36.69 & 38.52    \\
CoTexT          & 20.10 & 37.40 & 40.14    \\
UniXcoder       & 22.60 & 38.23 & -        \\
CodeT5-small    & 21.55 & 38.13 & 41.39    \\
CodeT5-base     & 22.30 & 40.73 & 43.20    \\
\hline
CodeT5-large    & \textbf{22.65} & \textbf{42.66} & \textbf{45.08}   \\
\hline
\end{tabular}

\vspace{+0.2in}

\centering
\caption{Code-to-Code generation results on CodeXGLUE}
\label{app_tab:codexglue_code2code}
\resizebox{1.0\textwidth}{!} {
\begin{tabular}{l| cc | c c  | c c | c c}
\hline
\multirow{2}{*}{Model} & \multicolumn{2}{c|}{Java to C\#} & \multicolumn{2}{c|}{C\# to Java}  & \multicolumn{2}{c|}{Refine Small}  & \multicolumn{2}{c}{Refine Medium}\\ 
\cline{2-9}
& BLEU-4 & EM  & BLEU-4 & EM & BLEU-4 & EM & BLEU-4 & EM  \\ 
\hline 
Naive copy & 18.54 & 0.00 & 18.69 & 0.00 & 78.06 & 0.00 & 90.91 & 0.00 \\ 
        Roborta (code) & 77.46 & 56.10 & 71.99 & 57.90 & 77.30 & 15.90 & 90.07 & 4.10 \\ 
        CodeBERT & 79.92 & 59.00 & 72.14 & 58.00 & 77.42 & 16.40 & 91.07 & 5.20 \\ 
        GraphCodeBERT & 80.58 & 59.40 & 72.64 & 58.80 & \textbf{80.02} & 17.30 & \textbf{91.31} & 9.10 \\ 
        PLBART & 83.02 & 64.60 & 78.35 & 65.00 & 77.02 & 19.21 & 88.50 & 8.98 \\ 
        CoTexT & - & - & - & - & 77.79 & 21.03 & 88.40 & 13.11 \\ 
        NSEdit & - & - & - & - & 71.06 & \textbf{24.04} & 85.72 & 13.87 \\ 
        CodeT5-small & 82.98 & 64.10 & 79.10 & 65.60 & 76.23 & 19.06 & 89.20 & 10.92 \\ 
        CodeT5-base & \textbf{84.03} & 65.90 & \textbf{79.87} & 66.90 & 77.43 & 21.61 & 87.64 & 13.96 \\ 
\hline        
        CodeT5-large & 83.56 & \textbf{66.00} & 79.77 & \textbf{67.00} & 77.38 & 21.70 & 89.22 & \textbf{14.76} \\ 
\hline
\end{tabular}
}

\end{table}
\begin{table}[t]
\centering
\caption{Ablation results of CodeRL with different CodeT5 model variants with different sizes, pretraining data and objectives on MBPP. CodeT5$^{\dag}$ is finetuned on APPS and evaluated on MBPP in a zero-shot setting.}\label{tab:mbpp}
\begin{tabular}{cc|cc|cc}
\hline
Model & Size & Data & Objective & \emph{pass@80} & \emph{pass@1000} \\    
\hline
\multicolumn{6}{c}{\cellcolor[HTML]{EFEFEF} GPT finetuned results} \\
\hline
GPT & 224M & Web Doc   & LM     & 7.2 &- \\ 
GPT & 422M & Web Doc   & LM     & 12.6 &-\\ 
GPT & 1B & Web Doc   & LM   & 22.4 &-\\       
GPT & 4B & Web Doc   & LM     & 33.0 &-\\   
GPT & 8B & Web Doc   & LM     & 40.6 &-\\   
GPT & 68B & Web Doc   & LM    & 53.6 &-\\   
GPT & 137B & Web Doc   & LM      & 61.4 &-\\   
\hline
\multicolumn{6}{c}{\cellcolor[HTML]{EFEFEF} CodeT5 finetuned results} \\
\hline
  
CodeT5 & 60M     & CSN       & MSP          & 19.2 & 36.2\\
CodeT5 &220M      & CSN     & MSP         & 24.0 &42.8\\
CodeT5 &770M        & CSN       & MSP         & 32.4& 47.8\\
CodeT5 &770M   & +GCPY    & MSP         & 34.6        &51.6\\
CodeT5 &770M       & +GCPY     & +NTP      & 46.8 &66.2\\
\hline
\multicolumn{6}{c}{\cellcolor[HTML]{EFEFEF} CodeRL zero-shot results} \\
\hline
CodeT5$^{\dag}$ &770M       & +GCPY      & +NTP       & 60.2&78.4\\
+CodeRL &770M       & +GCPY      & +NTP       & 63.0 &81.8\\
\hline
\end{tabular}
\end{table}

Table~\ref{tab:mbpp} reports the  \emph{pass@80} and  \emph{pass@1000} results for both finetuned and zero-shot settings.  For baselines, we compared with GPT models with sizes ranging from 224M to 137B~\citep{austin2021program}, which are pretrained on 2.93B web documents (13.8M containing source code) using standard language modeling objective.  Results of GPT models are obtained from the original authors.
From the comparison among various CodeT5 variants, we again confirm  that larger model sizes and pretraining data, and better pretraining objective of NTP all lead to a performance boost. Particularly, our CodeT5-770M yields a \emph{pass@80} of 46.8\%, surpassing GPT-8B's 40.6\% with a much smaller model size.
In addition, we find  CodeT5 models finetuned on APPS can achieve a surprisingly good zero-shot performance on MBPP with a \emph{pass@80} of 60.2\% and further improved to 63.0\% with the help of CodeRL, which even outperforms the largest GPT-137B's performance of 61.4\%.
This indicates APPS is a comprehensive  program synthesis benchmark and  CodeT5+CodeRL models trained on it are able to generalize to other simpler coding tasks.
If we further increase the budget of attempts up to 1000,  all models witness a consistent and significant boost in solving rate, especially our  CodeT5+CodeRL  yielding a new SOTA result of 81.8\% \emph{pass@1000}.

\section{Additional Qualitative Analysis}
\subsection{Failure Analysis} 
\label{app_sec:failure_analysis}

\begin{figure}[t]
	\centering
	\resizebox{1.0\textwidth}{!} {
	\includegraphics{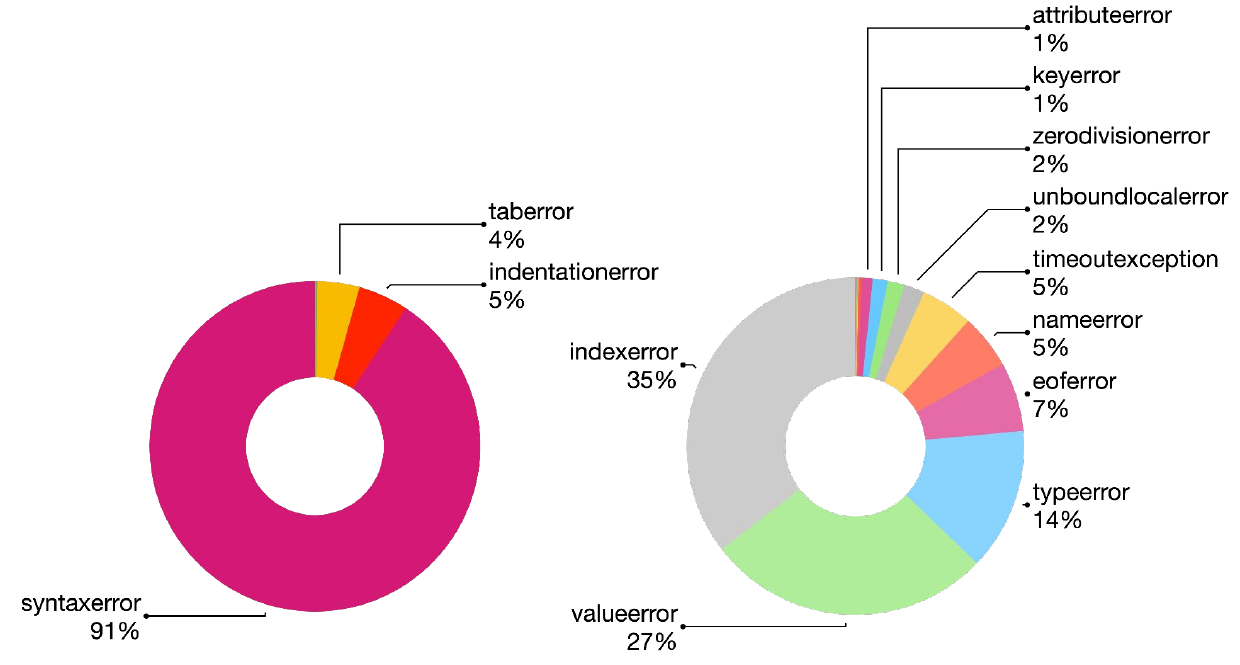}
	}
	\caption{
	\textbf{Compiling Errors (Left):}
	While current models are likely to generate programs with correct indentations, there are still more than $90\%$ of compiling errors due to syntactical mistakes.
	\textbf{Runtime Errors (Right):}
	Major runtime errors include \emph{indexerror, valueerror,} and \emph{typerror}. 
	Many of these errors occur due to mistakes in processing test inputs, e.g. wrong data types or mismatched numbers of elements. 
	Refer to Table \ref{app_tab:error_definition} for definitions of error types.
	}
	\label{app_fig:failure_analysis}
\end{figure}

Using a CodeT5+CodeRL model, we generate $200$ programs per sample in the APPS test splits.
We pass each program to the corresponding hidden unit tests.
We filter for samples with either runtime or compiling errors and extract the error types from the compiler.
From Figure \ref{app_fig:failure_analysis}-left, (and error definitions in Table \ref{app_tab:error_definition}), we observe that current models are able to probably indent lines of code, with only $4\%$ problems related to wrong \emph{tab} tokens and $5\%$ with wrong indentation levels. 
The majority of mistakes are syntactical problems, assuming more than $90\%$ of compiling errors. 

From Figure \ref{app_fig:failure_analysis}-right, among runtime errors, the most popular types of errors are due to wrong data index processing, inappropriate values, or mismatched data types. 
We found that many of these problems occur during preprocessing of test inputs, suggesting potential ways to improve current models in understanding and constructing proper input variables.

\begin{table}[t]
\centering
\small
\caption{
\textbf{Definitions of error types:}
Error definitions are extracted from official Python online documentation at \url{https://docs.python.org/3/tutorial/errors.html}.
}
\label{app_tab:error_definition}
\begin{tabular}{p{0.2\textwidth}p{0.7\textwidth}}
\hline
\textbf{Error Type}        & \multicolumn{1}{c}{\textbf{Description}}                                                                                                                            \\
\hline
\multicolumn{2}{l}{\textbf{Compiling Errors}}                                                                                                                                            \\
taberror          & Raised when indentation contains an inconsistent use of tabs and spaces. This is a subclass of IndentationError.                                           \\
indentationerror  & Base class for syntax errors related to incorrect indentation. This is a subclass of SyntaxError.                                                          \\
syntaxerror       & Raised when the parser encounters a syntax error. This may occur in an import statement, in a call to the built-in functions compile(), exec(), or eval(). \\
\hline
\multicolumn{2}{l}{\textbf{Runtime Errors}}                                                                                                                                              \\
attributeerror    & Raised when an attribute reference or assignment fails.                                                                                                    \\
keyerror          & Raised when a mapping (dictionary) key is not found in the set of existing keys.                                                                           \\
zerodivisionerror & Raised when the second argument of a division or modulo operation is zero.                                                                                 \\
unboundlocalerror & Raised when a reference is made to a local variable in a function or method, but no value has been bound to that variable.                                 \\
timeoutexception  & Raised when a system function timed out at the system level.                                                                                               \\
nameerror         & Raised when a local or global name is not found.                                                                                                           \\
eoferror          & Raised when the input() function hits an end-of-file condition (EOF) without reading any data.                                                             \\
typeerror         & Raised when an operation or function is applied to an object of inappropriate type.                                                                        \\
valueerror        & Raised when an operation or function receives an argument that has the right type but an inappropriate value.                                              \\
indexerror        & Raised when a sequence subscript is out of range.                                                                           \\
\hline
\end{tabular}
\end{table}

\subsection{Example Generated Programs} 
\label{app_sec:example_programs}

\begin{figure}[htbp]
	\centering
	\resizebox{1.0\textwidth}{!} {
	\includegraphics{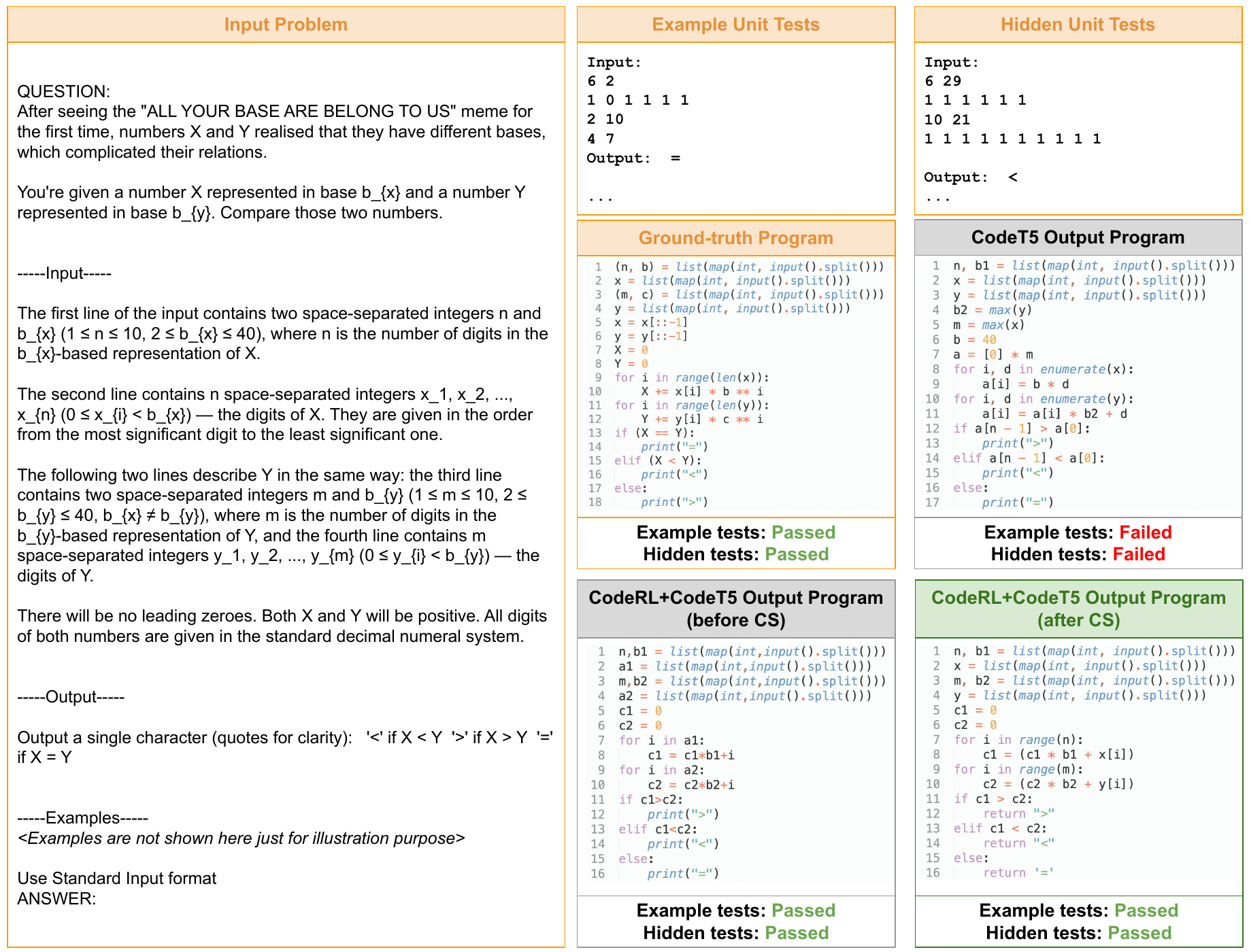}
	}
	\caption{
	\textbf{An example synthesis task from the APPS benchmark and corresponding programs generated by CodeT5 variants:}
	CodeRL+CodeT5 model can generate programs that pass both example tests and hidden tests, with or without the CS generation procedure. 
	}
	\label{app_fig:example_program4}
\end{figure}

\begin{figure}[htbp]
	\centering
	\resizebox{1.0\textwidth}{!} {
	\includegraphics{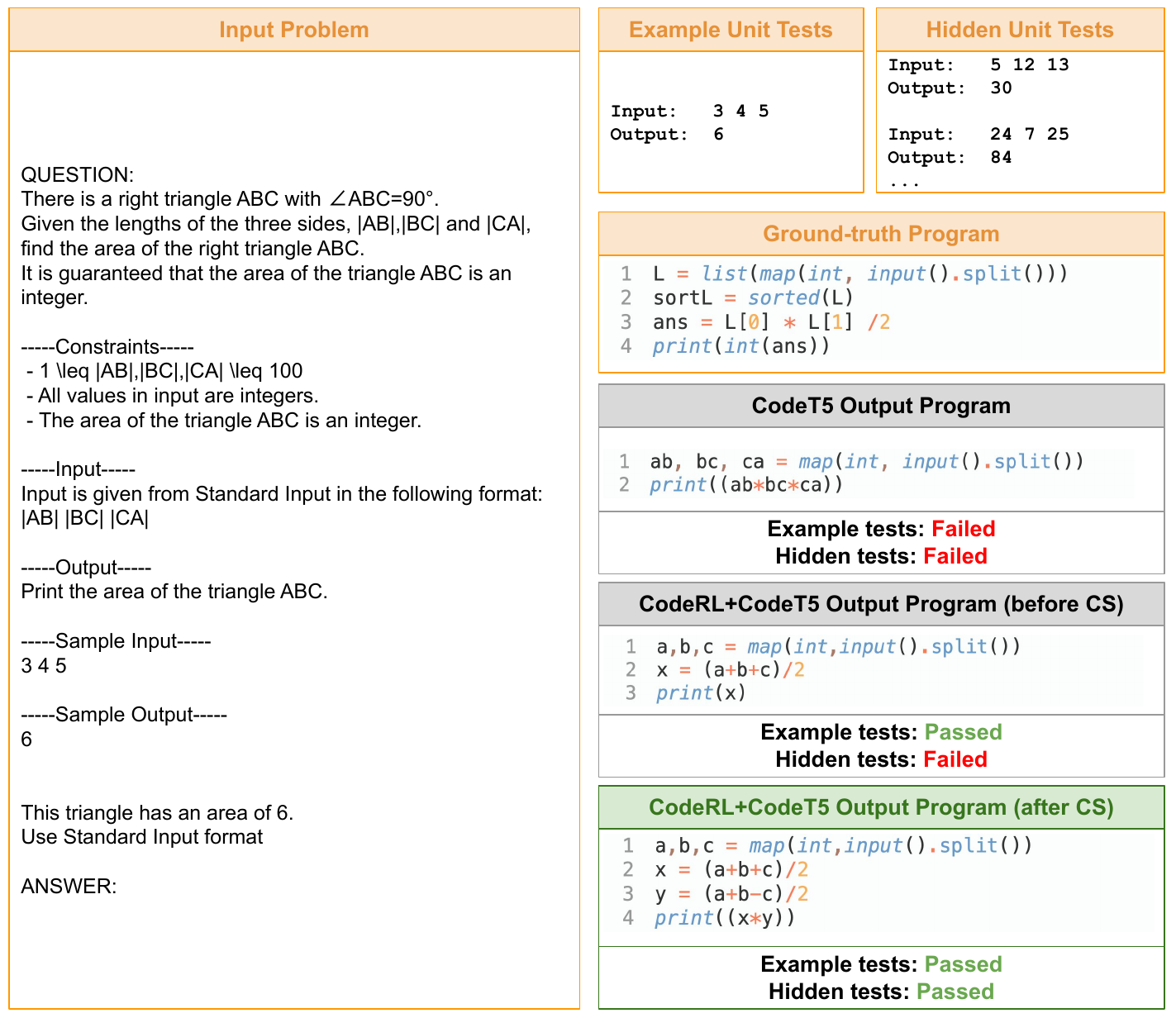}
	}
	\caption{
	\textbf{An example synthesis task from the APPS benchmark and corresponding programs generated by CodeT5 variants:}
	Without the CS generation procedure, CodeRL+CodeT5 model can generate programs that pass all example tests but fail hidden tests. 
	With the CS generation procedure, the model can condition on prior programs and generate a better program that passes all hidden tests. 
	}
	\label{app_fig:example_program1}
\end{figure}

\begin{figure}[htbp]
	\centering
	\resizebox{1.0\textwidth}{!} {
	\includegraphics{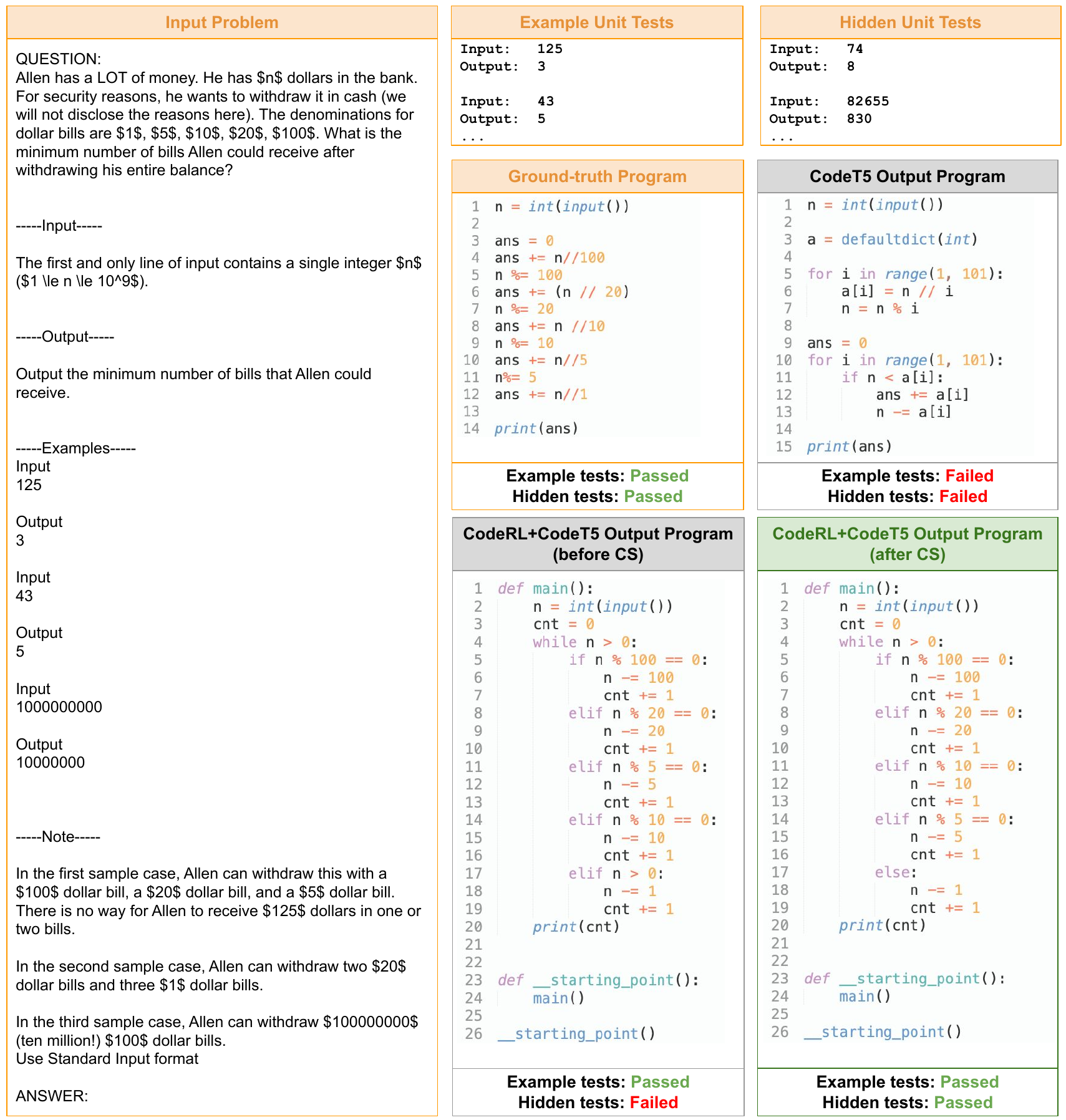}
	}
	\caption{
	\textbf{An example synthesis task from the APPS benchmark and corresponding programs generated by CodeT5 variants:}
	Without the CS generation procedure, CodeRL+CodeT5 model can generate programs that pass all example tests but fail hidden tests, especially those of corner cases. 
	With the CS generation procedure, the model can condition on prior programs and refine the code. 
	Specifically, we observe the model can simply reorder the \texttt{elif} blocks between line $11$ and $15$ to fix the error.
	The resulting program is functionally correct and passes all hidden tests. 
	}
	\label{app_fig:example_program3}
\end{figure}

\begin{figure}[htbp]
	\centering
	\resizebox{1.0\textwidth}{!} {
	\includegraphics{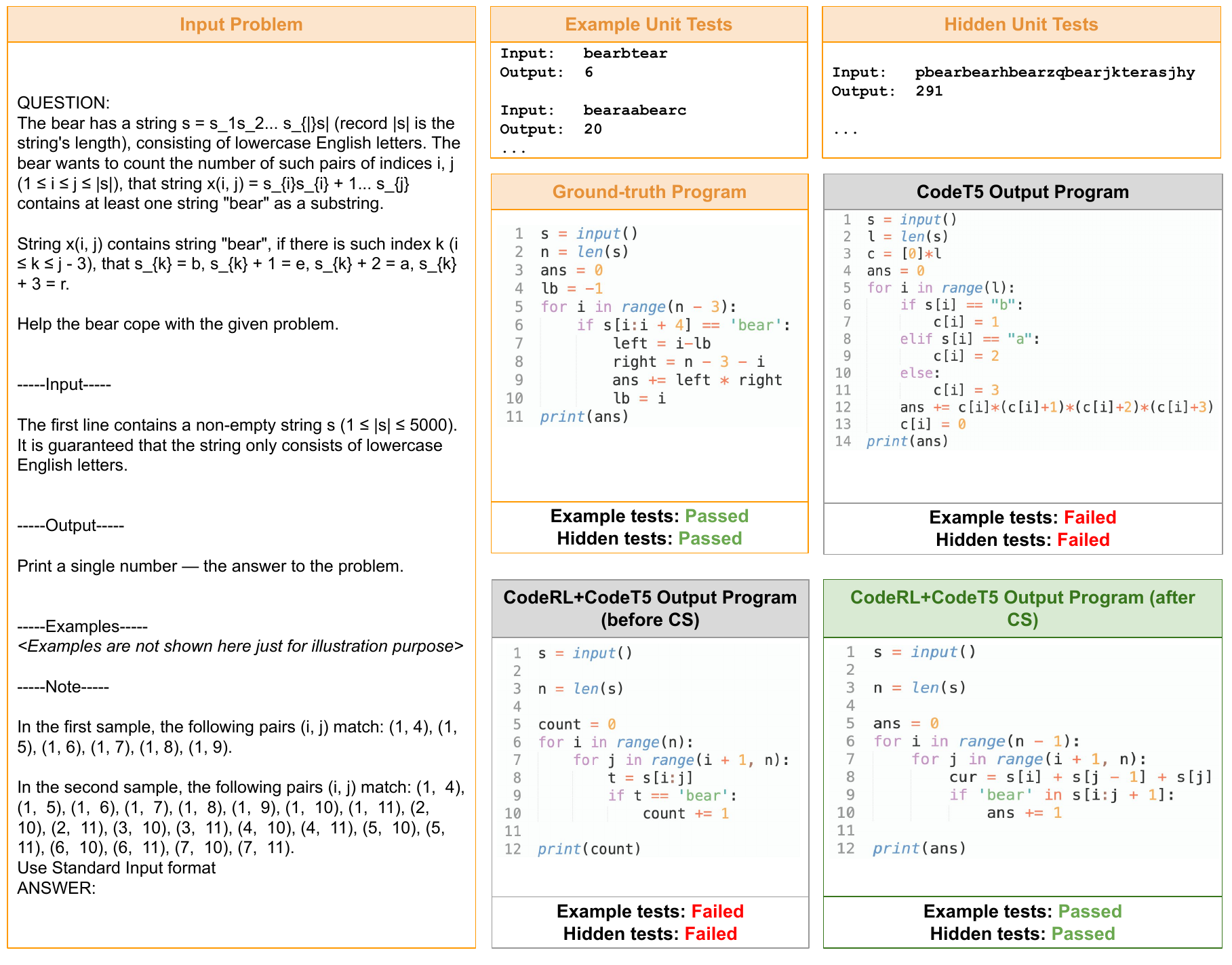}
	}
	\caption{
	\textbf{An example synthesis task from the APPS benchmark and corresponding programs generated by CodeT5 variants:}
	Without the CS generation procedure, CodeRL+CodeT5 model generates programs that fail example tests. 
	This scenario will trigger the CS generation procedure to firstly repair then refine the programs. 
	The resulting program can pass all hidden tests and fully satisfy the problem specification. 
	}
	\label{app_fig:example_program5}
\end{figure}

\begin{figure}[htbp]
	\centering
	\resizebox{1.0\textwidth}{!} {
	\includegraphics{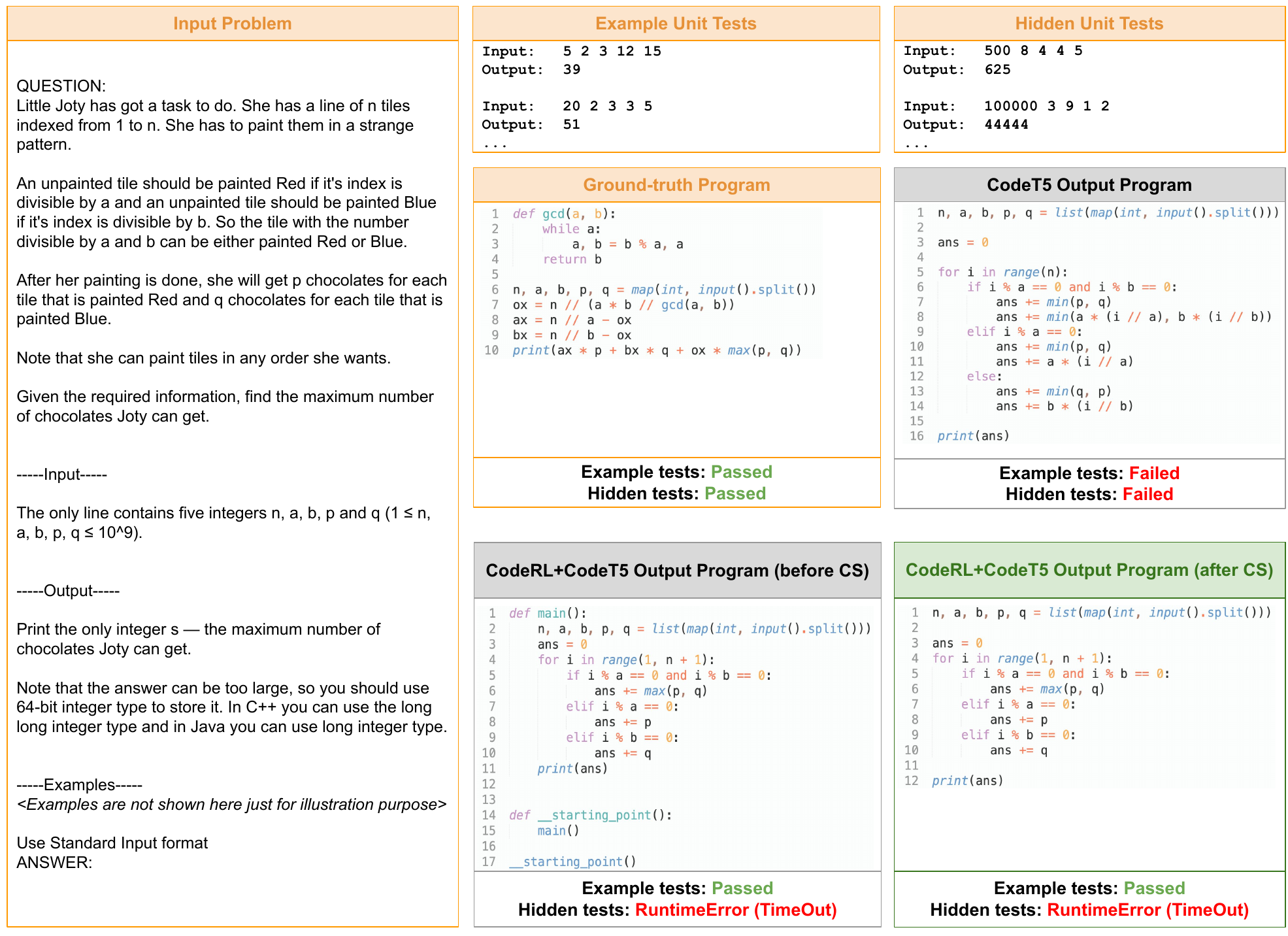}
	}
	\caption{
	\textbf{An example synthesis task from the APPS benchmark and corresponding programs generated by CodeT5 variants:}
	We demonstrate a failure case in which CodeRL+CodeT5 model generates incorrect programs, even with the application of the CS generation procedure. 
	Compared to CodeT5 model, applying CodeRL can improve the correctness of the programs but still fail during execution due to timeout errors. 
	The final program (after being refined by CS) still suffers from the same error and fails to pass hidden tests. 
	}
	\label{app_fig:example_program6}
\end{figure}

We present additional example generated programs in Figure \ref{app_fig:example_program4} to \ref{app_fig:example_program6}.
Specifically, we demonstrate cases where CodeRL+CodeT5 can successfully generate correct programs without the CS generation procedure (Figure \ref{app_fig:example_program4}), with CS via program refining (Figure \ref{app_fig:example_program1} and \ref{app_fig:example_program3}) and with CS via program repairing then refining (Figure \ref{app_fig:example_program5}).
In Figure \ref{app_fig:example_program6}, we demonstrate a failure case in which the final program still fails hidden tests. 
This failure example shows an opposite model behavior to the example in Figure \ref{app_fig:example_program2}, in which the CS generation procedure can successfully improve the efficiency of the output program to pass difficult test cases. 
As can be seen, compared to the ground-truth program, the output programs in Figure \ref{app_fig:example_program6} requires a lot more drastic modifications and it would be harder for the current CodeRL model to refine/regenerate the code. 

\end{document}